%% file: main.tex
\documentclass[a4paper, 10pt, conference]{ieeeconf}      
\usepackage{FG2026}

\FGfinalcopy 

\IEEEoverridecommandlockouts                              
\usepackage{graphics} 
\usepackage{amsmath} 
\usepackage{amssymb}  
\usepackage{booktabs}
\usepackage{multirow}
\usepackage{hyperref}
\usepackage{subcaption}
\usepackage{tikz}
\usepackage{url}

\usepackage{xspace}

\newcommand{\earlyexit}{\textit{EX-FIQA}\xspace}
\newcommand{\fusion}{\textit{EX-FIQA-F}\xspace}
\newcommand{\fusionw}{\textit{EX-FIQA-FW}\xspace}

\def\FGPaperID{59} 

\title{\LARGE \bf
EX-FIQA: Leveraging Intermediate Early eXit Representations from Vision Transformers for Face Image Quality Assessment
}

\author{\parbox{16cm}{\centering
    {\large Guray Ozgur$^{1,2}$, Tahar Chettaoui$^{1,2}$, Eduarda Caldeira$^{1,2}$, Jan Niklas Kolf$^{1,2}$, Andrea Atzori$^{1}$, Fadi Boutros$^{1}$, Naser Damer$^{1,2}$}\\
    {\normalsize
    $^1$ Fraunhofer Institute for Computer Graphics Research IGD, Darmstadt, Germany\\
    $^2$ Department of Computer Science, TU Darmstadt, Germany}
    \thanks{This research work has been funded by the German Federal Ministry of Education and Research and the Hessen State Ministry for Higher Education, Research and the Arts within their joint support of the National Research Center for Applied Cybersecurity ATHENE.}}
}

\begin{document}

\maketitle
\pagestyle{plain}
\thispagestyle{plain}

\begin{abstract}
Face Image Quality Assessment (FIQA) is crucial for reliable face recognition (FR) systems, yet existing Vision Transformer (ViT)-based approaches rely exclusively on final-layer representations, ignoring quality-relevant information captured at intermediate network depths. This paper presents the first comprehensive investigation of how intermediate representations within ViTs contribute to face quality assessment through early exit mechanisms and score fusion strategies. We systematically analyze all twelve transformer blocks of ViT-FIQA architectures, demonstrating that different depths capture distinct and complementary quality-relevant information, as evidenced by varying attention patterns and performance characteristics across network layers. Leveraging these insights, we propose a score fusion framework that combines quality predictions from multiple transformer blocks without architectural modifications or additional training. Our early exit analysis reveals optimal performance-efficiency trade-offs, enabling significant computational savings while maintaining competitive performance. Through extensive evaluation across eight benchmark datasets using four FR models, we demonstrate that our fusion strategy improves upon single-exit approaches. Our proposed quality fusion approach employs depth-weighted averaging that assigns progressively higher importance to deeper transformer blocks, achieving the best quality assessment performance by effectively leveraging the hierarchical nature of feature learning in ViTs. Our work challenges the conventional wisdom that only deep features matter for face analysis, revealing that intermediate representations contain valuable information for quality assessment. The proposed framework offers practical benefits for real-world biometric systems by enabling adaptive computation based on resource constraints while maintaining competitive quality assessment capabilities. The implementation is publicly available at:
\begin{footnotesize}
    \url{https://github.com/gurayozgur/EX-FIQA} 
\end{footnotesize}

\end{abstract}

\section{INTRODUCTION}
\label{sec:introduction}
FIQA measures \textit{recognition utility}, i.e. \textit{suitability for identity verification}, of facial images for recognition systems. Rather than evaluating aesthetic qualities such as sharpness or noise levels, FIQA specifically quantifies how effectively a facial image serves its purpose in verification scenarios, for example, determining whether images from passport scans or live captures contain sufficient biometric information for accurate identity matching \cite{NISTQuaity,DBLP:journals/csur/SchlettRHGFB22}. Unlike general Image Quality Assessment (IQA) methods that assess quality from human perception \cite{BRISQE_IQA, nique, liu2017rankiqa}, FIQA specifically evaluates face image suitability for automated recognition. As demonstrated by Biying et al. \cite{BiyingWACV}, high perceived quality does not always correlate with FR utility, particularly when factors like facial occlusions are present. Current state-of-the-art (SOTA) FIQA approaches primarily rely on representations from the final layer of deep networks, whether using convolutional neural networks (CNNs) \cite{PFE_FIQA, SERFIQ, MagFace,boutros_2023_crfiqa, grafiqs} or more recent ViT architectures \cite{atzori2025vitfiqaassessingfaceimage}. These approaches can be categorized into three groups: (1) \textit{supervised approaches} that train quality regressors from explicit or proxy labels \cite{faceqnetv1, SDDFIQA, best2018learning, RANKIQ_FIQA}, (2) \textit{unsupervised approaches} that estimate quality from fixed models without FIQA-specific supervision, including diffusion-based and FR-analysis methods \cite{10449044, babnikTBIOM2024, SERFIQ, grafiqs}, and (3) \textit{self-supervised (FR-integrated) approaches} that couple quality estimation with FR model learning \cite{boutros_2023_crfiqa, MagFace, PFE_FIQA, atzori2025vitfiqaassessingfaceimage}.

\begin{figure}[t]
    \centering
    \includegraphics[width=\linewidth]{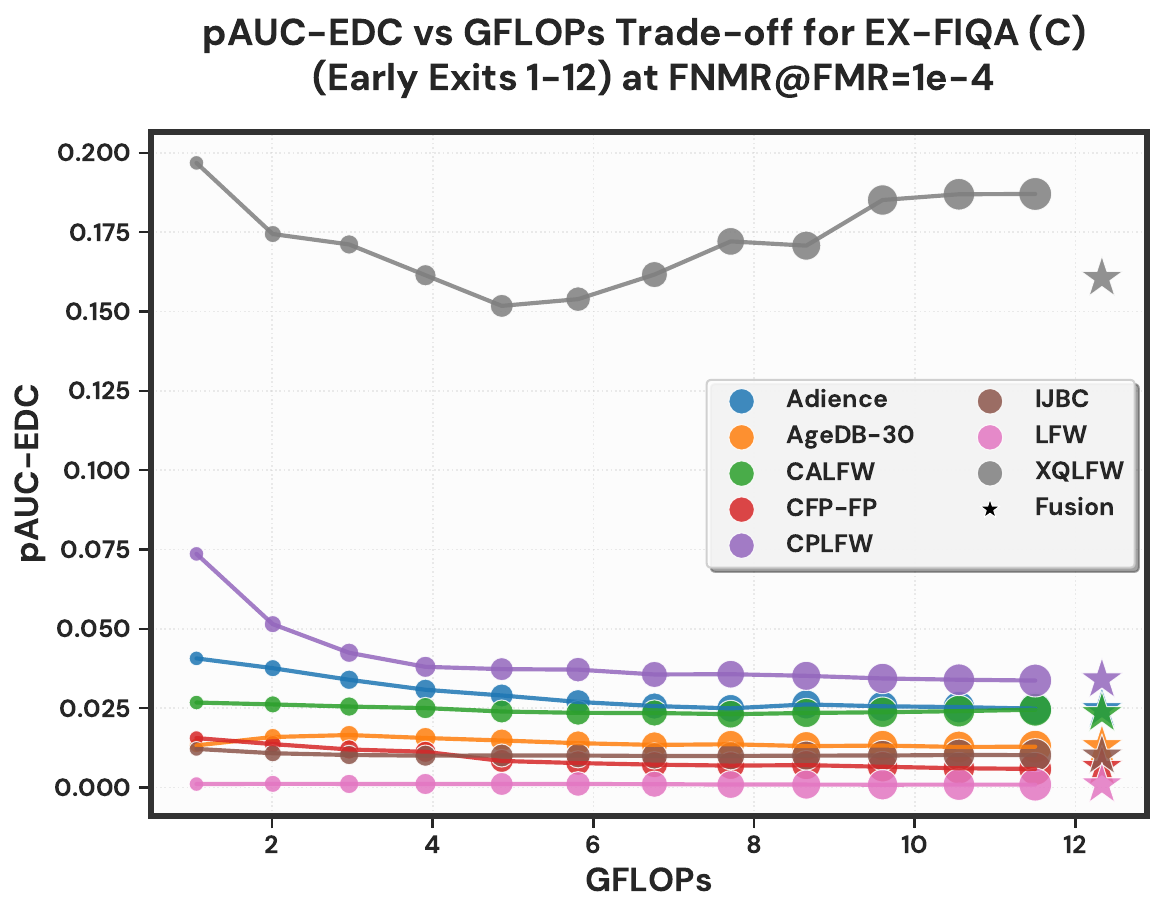}
    \caption{
        Performance of \earlyexit exits on FIQA across multiple datasets.
        The trade-off between computational complexity (GFLOPs) and performance (pAUC-EDC for FNMR@FMR =$1e-4$) is shown. Deeper exits generally improve pAUC-EDC (the lower the better), while incurring higher computational cost.
    }
    \label{fig:fnmr_vs_gflops}
\end{figure}

ViTs have recently gained significant attention in computer vision tasks, such as image classification \cite{liu2021swin}, object detection \cite{carion2020end,dai2021up}, semantic segmentation \cite{strudel2021segmenter,li2023focalunetr}, and image generation \cite{hudson2021generative}, and shown strong potential for FR, achieving comparable or superior performance through their attention-based architecture \cite{Dan_2023_TransFace, DBLP:conf/cvpr/KimS0JL24, DBLP:journals/ivc/ChettaouiDB25}. Unlike CNNs that process images through hierarchical local operations, ViTs divide images into patches and model global relationships through self-attention mechanisms \cite{lenet5,DBLP:conf/nips/VaswaniSPUJGKP17}. This architectural difference allows ViTs to capture long-range dependencies that may be particularly valuable for assessing face image quality, as demonstrated by recent approaches like ViT-FIQA \cite{atzori2025vitfiqaassessingfaceimage}, which extend standard ViT backbones with a learnable quality token designed to predict utility scores for face images. 

However, despite their performance advantages, ViTs are computationally expensive, making deployment on resource-constrained devices challenging \cite{DBLP:conf/iclr/DosovitskiyB0WZ21}. Early exits, a technique allowing inference to terminate at intermediate network depths, has emerged as a promising approach to address this limitation \cite{DBLP:conf/bmvc/BakhtiarniaZI21}. This approach is particularly valuable for surveillance applications, where frames with insufficient quality detected at early layers can be filtered out before proceeding to computationally expensive recognition stages, thereby decreasing the resource allocation in real-time systems. Additionally, ViTs process information hierarchically through transformer blocks, with earlier layers focusing on more generic visual features while deeper layers capture more task-specific representations \cite{DBLP:journals/corr/VaswaniSPUJGKP17}. This hierarchical processing suggests that quality-relevant information may be distributed across different network depths, rather than concentrated solely in the final layer. For FIQA, it remains unexplored whether earlier layers might capture important quality indicators that complement or even surpass the final layer's representation.

Motivated by these observations, we propose a comprehensive framework that leverages intermediate representations within ViTs to enhance face quality assessment. Our approach explores two key aspects:
\begin{itemize}
    \item We provide the first comprehensive analysis of early exits in ViTs for FIQA, quantifying performance-computation trade-offs across different network depths with \earlyexit.
    \item We propose \fusionw, a score fusion framework that leverages quality predictions from multiple transformer blocks to improve overall assessment accuracy with minimal computational overhead.
\end{itemize}

Our investigation spans all twelve transformer blocks of a ViT-S architecture by Atzori et al. \cite{atzori2025vitfiqaassessingfaceimage}. Through extensive experiments across eight benchmark datasets (Adience \cite{Adience}, AgeDB-30 \cite{agedb}, CFP-FP \cite{cfp-fp}, LFW \cite{LFWTech}, CALFW \cite{CALFW}, CPLFW \cite{CPLFWTech}, XQLFW \cite{XQLFW}, and IJB-C \cite{ijbc}), our approach demonstrates that middle-layer exits achieve competitive performance with up to 50\% computational savings while maintaining comparable performance to final-layer performance, illustrated in Figure \ref{fig:fnmr_vs_gflops}, and further discussed in Section \ref{sec:experimental_results}. Additionally, our \fusionw achieves better results on the most challenging large-scale benchmark, IJB-C \cite{ijbc}, with minimal computational overhead compared to SOTA, which will also be discussed in Section \ref{sec:experimental_results}. 

\begin{figure*}[ht]
\centering
\includegraphics[width=\textwidth, trim=1 1 1 1,clip]{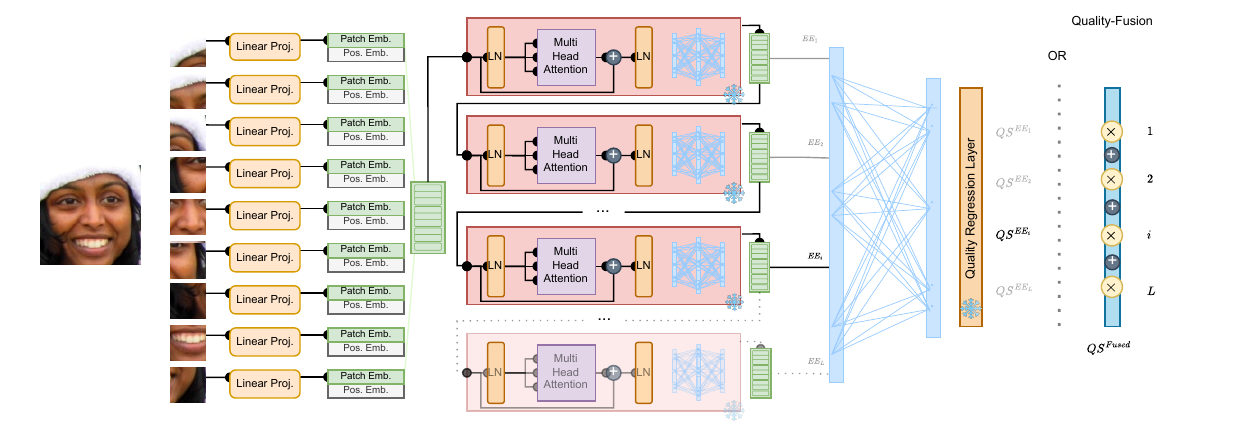}
\caption{Overview of \earlyexit (C) architecture and proposed early exit mechanism for inference. (1) A face image is divided into equally sized, non-overlapping patches and linearly projected to extract patch embeddings. (2) The patch embeddings are processed through a sequence of 12 transformer blocks. (3) Our early exit mechanism enables inference at any intermediate transformer block by extracting patch representations, concatenating them, and processing through the pre-trained two-layer feature network followed by the regression head to obtain quality scores. (4) The \fusionw framework combines predictions from multiple exit points using fusion strategies to achieve robust quality assessment. Early exits allow computational savings by terminating inference before processing all 12 transformer blocks.}
\label{fig:pipeline}
\end{figure*}

\section{RELATED WORK}
\label{sec:related_work}

\subsection{Face Image Quality Assessment}
\label{subsec:related_fiqa}

Building on our introduction, FIQA methods can be grouped into three paradigms. \textit{Supervised approaches} typically train quality regressors using explicit or proxy supervision. FaceQnet \cite{faceqnetv1} uses ICAO compliance standards as quality references, while SDD-FIQA \cite{SDDFIQA} employs distribution distances. RankIQ \cite{RANKIQ_FIQA} adopts a learning-to-rank strategy, training models to predict quality rankings based on FR performance metrics across different datasets. Subsequent works improve label reliability: CLIB-FIQA \cite{Ou_2024_CVPR} calibrates confidence of quality anchors, while MR-FIQA \cite{mrfiqa} leverages multi-reference representations from synthetic data to reduce label noise. \textit{Unsupervised approaches} can be divided into non-FR model methods and FR-specific methods. Non-FR model approaches estimate quality without conventional FIQA regressors or pre-trained FR models. DifFIQA \cite{10449044} leverages diffusion models to assess embedding robustness under different conditions, eDifFIQA \cite{babnikTBIOM2024} distills this into a lightweight predictor. FR-specific approaches probe frozen FR backbones without retraining to estimate FIQ. SER-FIQ \cite{SERFIQ} measures embedding stability under dropout perturbations, while GraFIQ \cite{grafiqs} uses gradient magnitudes during backpropagation to evaluate sample alignment with the FR model's objective. FaceQAN \cite{FaceQAN} estimates quality by quantifying adversarial robustness. ViTNT-FIQA \cite{Ozgur_2026_WACV} tracks embedding-trajectory stability across ViT layers, while FROQ \cite{froq} identifies informative intermediate layers via lightweight calibration to predict quality in a single forward pass. These methods leverage existing FR models but are constrained by their fixed representations.  \textit{Self-supervised approaches}, often implemented as FR-integrated methods, jointly optimize FR and FIQA. MagFace \cite{MagFace} links quality scores to embedding magnitudes through regularized training, while PFE \cite{PFE_FIQA} models embeddings as Gaussian distributions with uncertainty representing quality. CR-FIQA \cite{boutros_2023_crfiqa} estimates quality by predicting a sample's relative classifiability within the embedding space. ViT-FIQA \cite{atzori2025vitfiqaassessingfaceimage} extended standard ViT backbones with a learnable quality token designed to predict utility scores for face images, which has shown that ViTs can be effectively adapted for FIQA. Despite this progress, most methods still rely on final-layer representations, potentially missing quality-relevant signals available at intermediate depths. Our work addresses this gap by investigating transformer depth-wise representations, motivating early exits and fusion as practical mechanisms to balance efficiency and FIQA accuracy.

\subsection{Early Exits in ViTs}
\label{subsec:related_earlyexit}

ViTs process images through a sequence of self-attention blocks, each progressively refining the representation of the input \cite{DBLP:conf/iclr/DosovitskiyB0WZ21}. Prior work has shown that intermediate representations in ViTs capture complementary levels of abstraction: early layers focus on local low-level patterns (e.g., edges and textures), similar to the initial layers in CNNs; middle layers capture object parts and spatial relations; and deeper layers encode high-level semantic concepts useful for classification \cite{DBLP:conf/iclr/DosovitskiyB0WZ21, raghu2021vision}. This hierarchical refinement suggests that many inputs may become linearly separable well before reaching the final layer \cite{eesurvey3}. Early-exit mechanisms leverage this property by allowing inference to terminate at intermediate network depths, skipping the computation of remaining layers, and such dynamic inference approaches aim to reduce latency and energy consumption while preserving accuracy \cite{eesurvey3, eesurvey4}. ViTs are particularly well-suited to early exits because all transformer blocks output tokens of the same dimensionality, enabling the straightforward insertion of auxiliary classifiers after any block without the need for complex feature adaptation \cite{DBLP:conf/bmvc/BakhtiarniaZI21}. Moreover, since each block refines the same set of tokens, exiting early often provides usable predictions with minimal architectural modification. In this direction, Bakhtiarnia et al. \cite{DBLP:conf/bmvc/BakhtiarniaZI21} investigated early exits in ViTs for image classification, demonstrating that exits from later transformer blocks can achieve a favorable balance between accuracy and efficiency. In the context of biometrics, Nixon et al. \cite{earlyexitfr_biosig} explored a ViT architecture with early exits for face identification in a closed-set setting, showing that early exits after later transformer blocks can maintain recognition performance while reducing computational cost. Their work revealed that performance degradation increases dramatically for exits before block 9 in a 12-block architecture, suggesting that high-level descriptive facial features are achieved in middle-to-late transformer layers. Several complementary techniques have been proposed to further improve early-exit performance for ViTs, including specialized branch architectures with transformer encoders or convolutional heads  \cite{slvit}, Single-Layer ViT (SL-ViT) mechanisms that fuse local and global patch information \cite{slvit}, heterogeneous exit strategies like LGViT that employ different head designs for shallow versus deep layers \cite{lgvit}, self-distillation training approaches where branches mimic the final classifier's behavior \cite{distillation_multiexit}, and entropy-based dynamic routing policies that determine the optimal exit point based on prediction confidence \cite{BERxiT}. These methods enhance the discriminative power of intermediate classifiers and mitigate the accuracy gap between early and final exits. While such strategies could also benefit our task, which we leave it as future work, we focus here on evaluating early exits purely at inference time on a pre-trained FIQA model in order to investigate the effect of feature depth on quality assessment performance. Despite the growing interest in early exits for classification and recognition, their application to FIQA remains unexplored. This work addresses this gap by systematically analyzing how early exits affect FIQA predictions, providing insights into the depth at which quality-relevant features emerge in ViTs and enabling more computationally efficient quality assessment with minimal loss in accuracy.


\begin{figure}[htbp]
    \centering
    \includegraphics[width=0.45\textwidth]{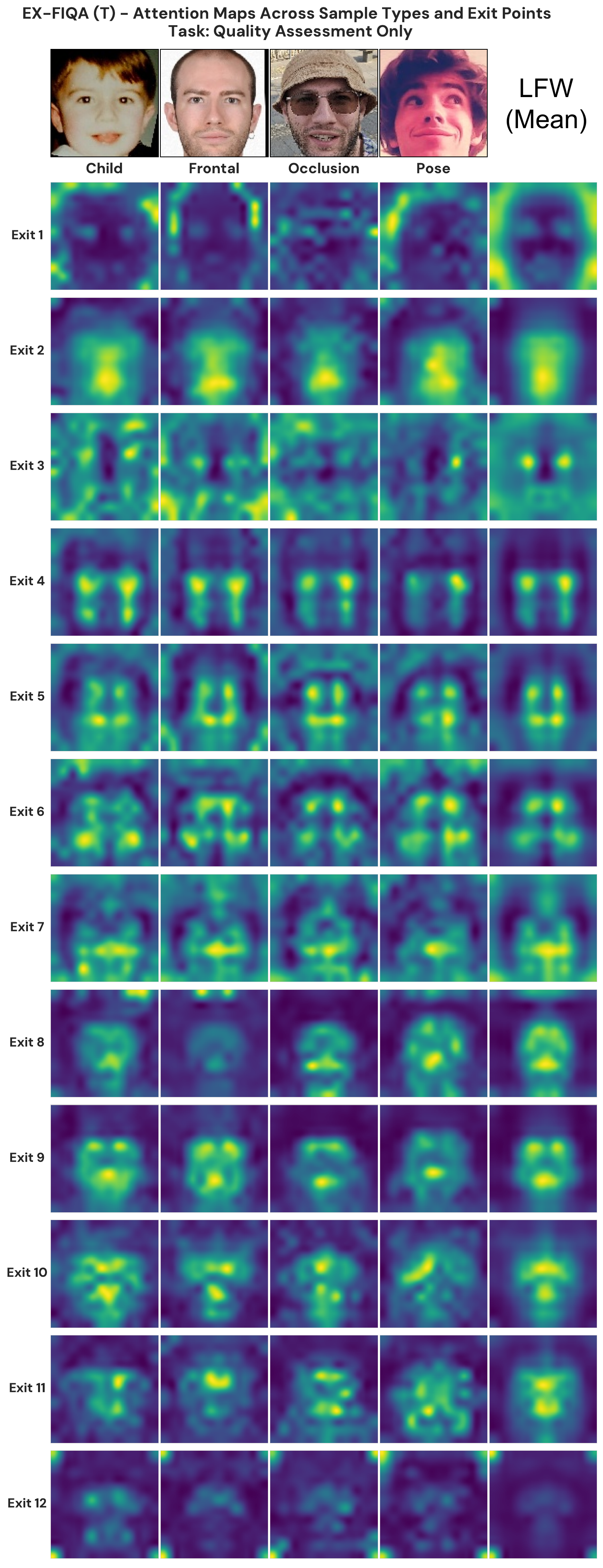}
    \caption{\earlyexit (T) attention map evolution of the quality token across different samples and exit points. Visualization of attention patterns for four representative face images: a child face, a frontal pose, an occluded face, and a challenging pose of the same person, and the mean of all images from LFW \cite{LFWTech} dataset. \textit{The individual appearing in the images have provided informed consent for their use in this research.}}
    \label{fig:token_axis4_sample}
\end{figure}

\section{METHODOLOGY}
\label{sec:methodology}
\subsection{Preliminary: ViT-FIQA}
\label{subsec:metho_vitfiqa}
Our work leverages pre-trained ViT-FIQA models from \cite{atzori2025vitfiqaassessingfaceimage}, utilizing two variants: ViT-FIQA (T) and ViT-FIQA (C) based on ViT. Figure \ref{fig:pipeline} provides an overview of the ViT-FIQA (C) architecture and illustrates our proposed early exit mechanism for inference and score fusion, where quality assessment can be performed at any intermediate transformer block by extracting representations and applying the pre-trained regression components or all transformer blocks could be used to get a fused score. We provide details on ViT-FIQA architectures, which enable our early exit and score fusion approaches on top of this architecture.

Both variants, ViT-FIQA (T) and ViT-FIQA (C), share core transformer processing components. Given a face image $\mathbf{x} \in \mathbb{R}^{H \times W \times C}$, it is divided into $N$ non-overlapping patches of size $P \times P$, where $N = \frac{HW}{P^2}$. Each patch is linearly projected to obtain patch embeddings $\mathbf{z}_0 \in \mathbb{R}^{N \times D}$.

The sequence is processed through $L=12$ transformer blocks, where each block $l$ applies the standard transformer architecture \cite{DBLP:conf/nips/VaswaniSPUJGKP17} with layer normalization (LN), multi-head self-attention (MSA), and multi-layer perceptron (MLP) components:
\begin{align}
\mathbf{Z}'_l &= \text{MSA}(\text{LN}(\mathbf{Z}_{l-1})) + \mathbf{Z}_{l-1} \\
\mathbf{Z}_l &= \text{MLP}(\text{LN}(\mathbf{Z}'_l)) + \mathbf{Z}'_l
\end{align}

The MSA mechanism with $H=8$ attention heads \cite{atzori2025vitfiqaassessingfaceimage} is defined as $\text{MSA}(\mathbf{X}) = \text{Concat}(\text{head}_1, \ldots, \text{head}_H) \mathbf{W}^O$, where each attention head is computed as $\text{head}_h = \text{Attention}(\mathbf{X}\mathbf{W}_h^Q, \mathbf{X}\mathbf{W}_h^K, \mathbf{X}\mathbf{W}_h^V)$ with projection matrices $\mathbf{W}_h^Q, \mathbf{W}_h^K, \mathbf{W}_h^V \in \mathbb{R}^{D \times d_h}$ and $d_h = D/H = 64$. The MLP implements a two-layer feed-forward network: $\text{MLP}(\mathbf{X}) = \text{ReLU6}(\mathbf{X}\mathbf{W}_1 + \mathbf{b}_1)\mathbf{W}_2 + \mathbf{b}_2$ with $\mathbf{W}_1 \in \mathbb{R}^{D \times 4D}$ and $\mathbf{W}_2 \in \mathbb{R}^{4D \times D}$.

\textbf{ViT-FIQA (T)} \cite{atzori2025vitfiqaassessingfaceimage} extends the conventional ViT with a learnable quality token $\mathbf{q}_0 \in \mathbb{R}^{1 \times D}$ concatenated to input patch embeddings. The complete input representation with position embeddings $\mathbf{E}_{pos} \in \mathbb{R}^{(N+1) \times D}$ is:
\begin{equation}
\mathbf{Z}_0 = [\mathbf{q}_0; \mathbf{z}_0] + \mathbf{E}_{pos}
\end{equation}

resulting in a $(N+1) \times D$ matrix. The quality score is computed from the final quality token through a batch-normalized regression head:
\begin{equation}
s = \mathbf{W}_T \text{BN}(\mathbf{q}_L)
\end{equation}

where $\mathbf{W}_T \in \mathbb{R}^{1 \times D}$ is the regression weight matrix without bias term.

\textbf{ViT-FIQA (C)} \cite{atzori2025vitfiqaassessingfaceimage} follows the conventional CR-FIQA approach \cite{boutros_2023_crfiqa}, processing only patch embeddings without additional tokens. The input sequence consists solely of patch embeddings with position encodings: $\mathbf{Z}_0 = \mathbf{z}_0 + \mathbf{E}_{pos}$, where $\mathbf{E}_{pos} \in \mathbb{R}^{N \times D}$ are learnable position embeddings.

After transformer processing, patch representations are concatenated and processed through a two-layer feature network:
\begin{align}
\mathbf{h} &= [\mathbf{z}_{L,1}; \mathbf{z}_{L,2}; \ldots; \mathbf{z}_{L,N}] \in \mathbb{R}^{N \times D} \\
\mathbf{f}_1 &= \text{BN}(\mathbf{W}_{fc1} \mathbf{h}_{flat}) \\
\mathbf{f} &= \text{BN}(\mathbf{W}_{fc2} \mathbf{f}_1)
\end{align}

where $\mathbf{h}_{flat} \in \mathbb{R}^{N \cdot D}$ is the flattened concatenation, $\mathbf{W}_{fc1} \in \mathbb{R}^{D \times (N \cdot D)}$ and $\mathbf{W}_{fc2} \in \mathbb{R}^{D \times D}$ are weight matrices without bias terms. The quality score is computed as: $s = \mathbf{W}_C \text{BN}(\mathbf{f})$ where $\mathbf{W}_C \in \mathbb{R}^{1 \times D}$.

Both models were trained by Atzori et al. \cite{atzori2025vitfiqaassessingfaceimage} with combined objectives using CR-FIQA loss \cite{boutros_2023_crfiqa}: $\mathcal{L} = \mathcal{L}_{FR} + \lambda \mathcal{L}_{FIQ}$, where $\mathcal{L}_{FR}$ is the CosFace margin penalty softmax loss \cite{DBLP:conf/cvpr/WangWZJGZL018} and $\mathcal{L}_{FIQ}$ is Smooth L1 regression loss between predicted quality scores and CR-FIQA target values with $\lambda=10$ \cite{atzori2025vitfiqaassessingfaceimage}.

\subsection{Early Exit Mechanism: \earlyexit}
\label{subsec:metho_earlyexit}
We implement early exits by extracting intermediate representations from each transformer block $l \in \{1, 2, \ldots, L\}$ and applying the original pre-trained quality assessment components without introducing additional parameters. The key insight underlying our early exit approach stems from a fundamental architectural difference between ViTs and CNNs. While CNNs progressively reduce spatial dimensions while increasing channel depth (e.g., from $224 \times 224 \times 3$ input to $112 \times 112 \times 64$, then $56 \times 56 \times 128$), ViTs maintain constant token dimensionality $D=512$ throughout all $L=12$ transformer layers. This architectural property enables three critical advantages for early exit implementation: (1) \textbf{Dimensional compatibility}: Every transformer block outputs representations in $\mathbb{R}^{N \times D}$ format (where $N=144$ patches), allowing direct application of pre-trained regression heads $\mathbf{W}_T \in \mathbb{R}^{1 \times D}$ or $\mathbf{W}_C \in \mathbb{R}^{1 \times D}$ to any intermediate layer without architectural modifications \cite{DBLP:journals/corr/VaswaniSPUJGKP17}. (2) \textbf{Parameter reusability}: Each transformer block refines embeddings within the same $D$-dimensional space \cite{DBLP:conf/iclr/DosovitskiyB0WZ21}, enabling the original quality assessment parameters trained on final-layer representations to be directly applied to intermediate representations from any block $l \in \{1, 2, \ldots, L\}$. (3) \textbf{Structural consistency}: Unlike CNNs requiring complex feature adaptation layers for early exits \cite{DBLP:conf/icpr/Teerapittayanon16}, ViTs enable seamless prediction at any depth using identical computational pathways \cite{DBLP:conf/icml/TouvronCDMSJ21}.

For ViT-FIQA (T) \cite{atzori2025vitfiqaassessingfaceimage}, we extract the quality token $\mathbf{q}_l \in \mathbb{R}^D$ from transformer block $l$ and compute the quality score using the original pre-trained regression head: $s_l^{(T)} = \mathbf{W}_T \text{BN}(\mathbf{q}_l)$, where $\mathbf{W}_T \in \mathbb{R}^{1 \times D}$ is the learned regression weight matrix from the original ViT-FIQA (T) \cite{atzori2025vitfiqaassessingfaceimage} training that projects the batch-normalized $D$-dimensional quality token to a scalar quality score. For ViT-FIQA (C) \cite{atzori2025vitfiqaassessingfaceimage}, we concatenate patch representations $\mathbf{z}_{l,i}$ from transformer block $l$ to form $\mathbf{h}_l = [\mathbf{z}_{l,1}; \mathbf{z}_{l,2}; \ldots; \mathbf{z}_{l,N}] \in \mathbb{R}^{N \times D}$, apply the pre-trained two-layer feature network to obtain intermediate face embedding $\mathbf{f}_l \in \mathbb{R}^D$, then compute: $s_l^{(C)} = \mathbf{W}_C \text{BN}(\mathbf{f}_l)$, where $\mathbf{W}_C \in \mathbb{R}^{1 \times D}$ is the original regression weight matrix from ViT-FIQA (C) training. Both $\mathbf{W}_T$ and $\mathbf{W}_C$ are pre-trained parameters that can be directly applied to intermediate representations due to the constant $D$-dimensional structure across all transformer blocks. From here on, we name \earlyexit (C)/(T) for our early exit implementation of ViT-FIQA (C)/(T) respectively.

\textbf{Computational Complexity}
Early exit at block $l$ provides computational savings by skipping $(L-l)$ transformer blocks. However, quality assessment overhead significantly impacts actual savings. \earlyexit (T) requirees only $\approx$3000 FLOPs per exit (batch normalization + regression), while \earlyexit (C) requires $\approx$76M FLOPs per exit due to the two-layer feature network processing concatenated patches. This overhead partially limits computational reduction, especially for \earlyexit (C).

\subsection{Score Fusion Strategies: \fusion \& \fusionw}
\label{subsec:metho_fusion}
While early exits enable computational savings, they may sacrifice accuracy in some settings, especially when exiting from very early blocks, compared to using the full network depth. However, the intermediate representations capture distinct quality-relevant information that, when properly aggregated, can potentially exceed the performance achieved by relying solely on the final layer. This is grounded in our empirical observations from attention map analysis; see Figure \ref{fig:token_axis4_sample}: Different intermediate layers capture distinct levels of semantic information, with early blocks focusing on low-level features while deeper blocks encode higher-level attributes. Score fusion enables complementary information to be collected by leveraging the distributed quality information across multiple network depths. Let $\mathbf{s} = [s_1, s_2, \ldots, s_L]^T$ denote the vector of quality scores from all $L$ transformer blocks. Our fusion strategies combine these predictions using weighted combinations: $s_{\text{fused}} = \sum_{l=1}^{L} w_l \cdot s_l$, where different weight vectors $\mathbf{w}$ define distinct fusion approaches.

We propose two fusion strategies: (1) \textbf{\fusion} implements uniform weighting ($w_l = 1/L$) across all transformer blocks, assuming quality-relevant information is uniformly distributed across network depths. (2) \textbf{\fusionw} implements depth-based linear weighting ($w_l \propto l$), assigning progressively higher importance to deeper blocks based on the assumption that representational quality improves with depth, as we will show empirically in Table \ref{tab:ablation_pauc}, and visually in Figure \ref{fig:token_axis4_sample}, later stages achieve better results than earlier stages, which supports the assumption of improvement in representational quality with depth. The fusion strategies require processing through all transformer blocks but introduce additional computational overhead from applying quality assessment components at each intermediate layer. For \earlyexit (T), the overhead is minimal ($\approx$3000 FLOPs per exit), while \earlyexit (C) incurs substantial overhead ($\approx$76M FLOPs per exit) due to the two-layer feature network. 

\input{tab/ablation_pauc}

\section{EXPERIMENTAL DETAILS}
\label{sec:experimental_details}
\subsection{Implementation Details and Model Setup}
We utilize the pre-trained ViT-FIQA models \cite{atzori2025vitfiqaassessingfaceimage} without modification in the base architecture, extracting intermediate representations from each transformer intermediate block and applying the original pre-trained regression components. This ensures our analysis reflects the true capabilities of pre-trained representations.

\subsection{Datasets and Evaluation Protocols}
We conducted extensive experiments across eight benchmark datasets: LFW \cite{LFWTech}, AgeDB-30 \cite{agedb}, CFP-FP \cite{cfp-fp}, CALFW \cite{CALFW}, Adience \cite{Adience}, CPLFW \cite{CPLFWTech}, XQLFW \cite{XQLFW}, and IJB-C \cite{ijbc}. Performance was measured using Error-versus-Discard Characteristic (EDC) curves \cite{GT07}, which assess the impact of discarding low-quality face images on face verification performance and quantify how verification errors decrease as low-quality samples are progressively removed. The False Non-Match Rate (FNMR) was evaluated at fixed False Match Rate (FMR) thresholds \cite{iso_metric}, specifically at $1e-3$ (recommended for border control by Frontex \cite{frontex2015best}) and $1e-4$ (for higher security applications). Additionally, we reported the Area Under the Curve (AUC) and partial AUC (pAUC) of the EDC curves to quantify verification performance across rejection rates. The pAUC measures performance up to a 30\% rejection rate following established protocols \cite{10449044, babnikTBIOM2024, DBLP:journals/tbbis/SchlettRTB24}. To thoroughly examine the impact of our FIQA approaches across different FR architectures, we evaluated performance using four CNN-based models: ArcFace \cite{deng2019arcface}, ElasticFace \cite{elasticface}, MagFace \cite{MagFace}, and CurricularFace \cite{curricularFace}. All evaluations were conducted under cross-model settings, where the models used to learn FIQA were different from those used to extract face feature representations.


\input{tab/pAUC_all}

\section{EXPERIMENTAL RESULTS}
\label{sec:experimental_results}
\subsection{Early Exit and Fusion Performance Analysis}

Table \ref{tab:ablation_pauc} presents the pAUC-EDC results across eight benchmark datasets using ArcFace \cite{deng2019arcface} as the FR model, where lower pAUC-EDC values indicate superior quality assessment performance. The results are reported from different early exits (\earlyexit Exit 1-12) and score fusion strategies (\fusion and \fusionw) as described in Section \ref{sec:methodology}.

\textbf{Performance Evolution and Fusion Results:} For \earlyexit (C), we observe, in general, a performance improvement trajectory from early to deeper exits. The most significant performance gains occur after the first 4 layers, where the mean pAUC-EDC drops from 36.978/47.487 (Exit 1) to 27.255/34.536 (Exit 5). In comparison to the improvement from the first 4 exists, after the first 4 layers, the performance gain is limited, where we achieve a mean pAUC-EDC of 26.526/37.528 for the best exit (Exit 11). Although Exit 12 achieves some of the best individual performance on some small-scale benchmarks, such as CFP-FP and CPLFW, the best pAUC-EDC is achieved from the exits from middle layers for the challenging benchmarks. For instance, Exit 8 was the best performing on CALFW, and Exit 7 and 8 were the best performing ones on IJB-C. Similarly, \earlyexit (T) demonstrates performance improvements with increasing depth, though with a more gradual trajectory. The performance improvement from Exit 1 to Exit 12 shows a reduction in mean pAUC-EDC (from 35.187/45.154 to 26.481/33.294), with optimal mean performance achieved at Exit 10. Exit 12 exhibited the best pAUC-EDC values of 9.948/25.664 for Adience and 20.531/33.388 for CPLFW. For IJB-C, similar to \earlyexit (C), which exists from the middle layer (Exit 7-8-9), achieved the best results. Our score fusion results demonstrate the effectiveness of the \fusionw framework in combining predictions from multiple transformer depths, achieving better performance. The weighted averaging approach (\fusionw) consistently outperforms the final exit (Exit 12) and uniform averaging (\fusion) across both variants (C)/(T) for mean pAUC-EDC at FNMR@FMR=$1e-3$. Specifically, \fusionw (C) achieves a mean pAUC of 26.030/34.064, an improvement over Exit 12 (26.030/34.064), and surpasses Exit 12 in 5 out of 8 datasets. The weighted approach proves particularly effective on challenging datasets, achieving notable improvements on Adience (9.768/24.601  vs. 9.948/24.833 for Exit 12) and on IJB-C (6.503/9.915 vs. 6.599/10.185 for Exit 12) and maintaining strong performance across diverse scenarios.

\textbf{Computational Analysis for Early Exits and Fusion:} Computational costs are reported in Floating Point Operations (FLOPs), providing a hardware-agnostic metric to quantify efficiency improvements. A crucial finding from our analysis is that middle-layer exits (Exits 6-10) achieve competitive performance compared to the final layer while offering substantial computational savings, as shown in Table \ref{tab:ablation_pauc}. For \earlyexit (C), Exit 7 achieves a similar performance to Exit 12 while reducing computation (6.75 vs 11.49 GFLOPs). Similarly, Exit 11 performs better than Exit 12 with computational savings (10.55 vs 11.49 GFLOPs). For \earlyexit (T), Exit 10 improves upon Exit 12's mean performance (25.866 vs 26.481) while saving 17\% computation, and Exit 8 achieves comparable results with 33\% computational reduction. This demonstrates that the quality-relevant features necessary for robust face image quality assessment are largely captured already by the middle transformer blocks, making these exits highly attractive for deployment scenarios where computational efficiency is critical. The marginal performance gains achieved by processing through all 12 layers often do not justify the additional computational cost, particularly for real-time applications or resource-constrained environments, only for the last exit results. For fusion strategies, while they require processing through all transformer blocks, they introduce different computational overheads depending on the architecture. For \fusionw (T), the fusion overhead is minimal, as there is already a quality token in each block to be used in quality regression. For \fusionw (C), fusion incurs reasonable overhead, increasing total computation to 12.33 GFLOPs ($\approx$7\% overhead) from 11.49 GFLOPs.

\textbf{Operating Point Recommendations:} Based on our analysis, we identify several optimal operating points for different deployment scenarios: (1) \textbf{Maximum Efficiency}: \earlyexit (T) Exit 6 offers 50\% computational savings with minimal performance loss (26.481$\rightarrow$27.070 for FNMR@FMR=$1e-3$), ideal for resource-constrained applications. (2) \textbf{Balanced Performance}: \earlyexit (C) Exit 7 provides 41\% computational savings with acceptable performance trade-offs (26.664$\rightarrow$26.988 for FNMR@FMR=$1e-3$), suitable for real-time applications requiring good accuracy. (3) \textbf{Maximum Accuracy}: \fusionw for both variants (C)/(T) achieves optimal performance with minimal computational overhead for \earlyexit (T) and reasonable overhead for \earlyexit (C), recommended for high-accuracy requirements.

\subsection{Comparison with State-of-the-Art}
Table \ref{tab:sota_pauc} presents a comprehensive comparison of our approach with SOTA methods across four FR models, evaluating against three IQA methods \cite{BRISQE_IQA, liu2017rankiqa, DEEPIQ_IQA} and twelve FIQA approaches \cite{RANKIQ_FIQA, PFE_FIQA, SERFIQ, faceqnetv1, MagFace, SDDFIQA,boutros_2023_crfiqa, 10449044, babnikTBIOM2024, grafiqs, Ou_2024_CVPR, atzori2025vitfiqaassessingfaceimage}. Although our main goal is not to compete with SOTA, \fusionw (T) consistently achieves top-1 performance on the most challenging, large-scale IJB-C dataset across all FR models. On Adience, another large-scale dataset, \fusionw (C) achieves the best performance across all FR models, while achieving competitive results on small-scale evaluation benchmarks, including AgeDB-30, CFP-FP, LFW, CALFW, CPLFW, and XQLFW. The results highlight that our \fusionw framework effectively leverages intermediate transformer representations to outperform both traditional and recent deep learning-based FIQA approaches while maintaining computational efficiency for (T), or with a reasonable computational overhead for (C).


\section{CONCLUSION}
\label{sec:conclusion}
This paper presents the first comprehensive investigation of how intermediate representations within Vision Transformers contribute to Face Image Quality Assessment through early exit mechanisms and score fusion strategies, challenging the conventional wisdom that only final-layer representations matter for face image quality analysis. Our systematic analysis of all twelve transformer blocks in \earlyexit architectures reveals that different network depths capture distinct and quality-relevant information, with middle-layer exits (6-10) achieving up to 50\% computational reduction while maintaining competitive performance, particularly valuable for real-time biometric systems where computational efficiency is critical. We propose \fusionw, a novel score fusion framework that effectively combines predictions from multiple transformer depths using depth-weighted averaging, consistently outperforming uniform averaging and final-layer predictions across challenging benchmark datasets and achieving top-1 performance on large-scale datasets like IJB-C and Adience. Through extensive evaluation across eight benchmark datasets using four state-of-the-art face recognition models, we establish optimal operating points for different deployment scenarios: maximum efficiency (Exit 6 with 50\% computational savings), balanced performance (Exit 7 with 41\% savings), and maximum accuracy (\fusionw fusion). The practical implications extend beyond performance improvements by enabling adaptive computation based on resource constraints, addressing deployment challenges of ViTs in resource-constrained biometric systems, and opening new possibilities for edge computing applications in surveillance and access control systems.

\section*{ETHICAL IMPACT STATEMENT}
Our FIQA research aims to improve biometric system reliability by better identifying high-quality face images, potentially enhancing security and accessibility to identity verification services. However, FIQA systems present ethical risks, including biased quality assessments that may systematically disadvantage certain demographic groups, leading to discriminatory access to services, and enabling more effective mass surveillance through automated quality filtering of large image datasets. To mitigate these concerns, we emphasize training FIQA models on diverse datasets, implementing regular bias testing, and ensuring deployment within appropriate legal frameworks with proper oversight. We encourage the development of fair, transparent FIQA systems that serve all populations equitably and reject all malicious or illegal applications.


\clearpage
{\small
\bibliographystyle{ieee}
\bibliography{egbib}
}

\clearpage
\appendix

This supplementary material provides comprehensive experimental results and detailed analysis of the \earlyexit method for face image quality assessment. The supplementary material is structured to address four fundamental research questions: (1) How do quality predictions evolve across transformer exits? (2) Which exit points provide optimal performance? (3) How does our method compare against existing approaches? (4) What computational trade-offs exist?

\textbf{Tables: Quantitative Evidence}
\begin{itemize}
    \item \textbf{Table \ref{tab:quality_scores}}: Quality score evolution across transformer exits for representative face samples, showing how quality predictions change at each exit point for different face types (child, frontal, occlusion, pose variations). Higher scores indicate better predicted quality, with rankings in parentheses showing relative quality order for each exit. We include this detailed score evolution to demonstrate the progressive refinement. Quality predictions improve and stabilize as information flows through deeper transformer layers, providing concrete evidence that intermediate representations capture meaningful quality information at different abstraction levels, even as seen by the rankings.

    \item \textbf{Table \ref{tab:ablation_auc}}: Comprehensive ablation study showing AUC-EDC performance for all 12 individual exits and fusion strategies across 8 benchmark datasets using ArcFace as the face recognition model. Lower AUC-EDC values indicate better quality assessment performance. This systematic evaluation is essential to identify the optimal exit point and validate our fusion strategies, in addition to the pAUC-EDC table that we have in the main paper.

    \item \textbf{Table \ref{tab:sota_auc}}: State-of-the-art comparison presenting AUC-EDC values for our method against 15 competing approaches (3 IQA and 12 FIQA methods) across multiple face recognition models (ArcFace, ElasticFaceMagFace, and CurricularFace). We provide this extensive comparison to establish the competitive advantage of our approach, in addition to the pAUC-EDC table that we have in the main paper.

    \item \textbf{Table \ref{tab:ijbc_ver}}: Verification performance analysis on the IJB-C dataset showing True Acceptance Rate (TAR) at different False Acceptance Rate (FAR) thresholds for each exit and fusion strategy across six face recognition models by using quality score instead of faceness score. This verification analysis serves to validate the practical utility of our quality assessments in real-world face recognition scenarios.
    
\end{itemize}

\begin{figure}[h]
    \centering
    \includegraphics[width=0.4\textwidth]{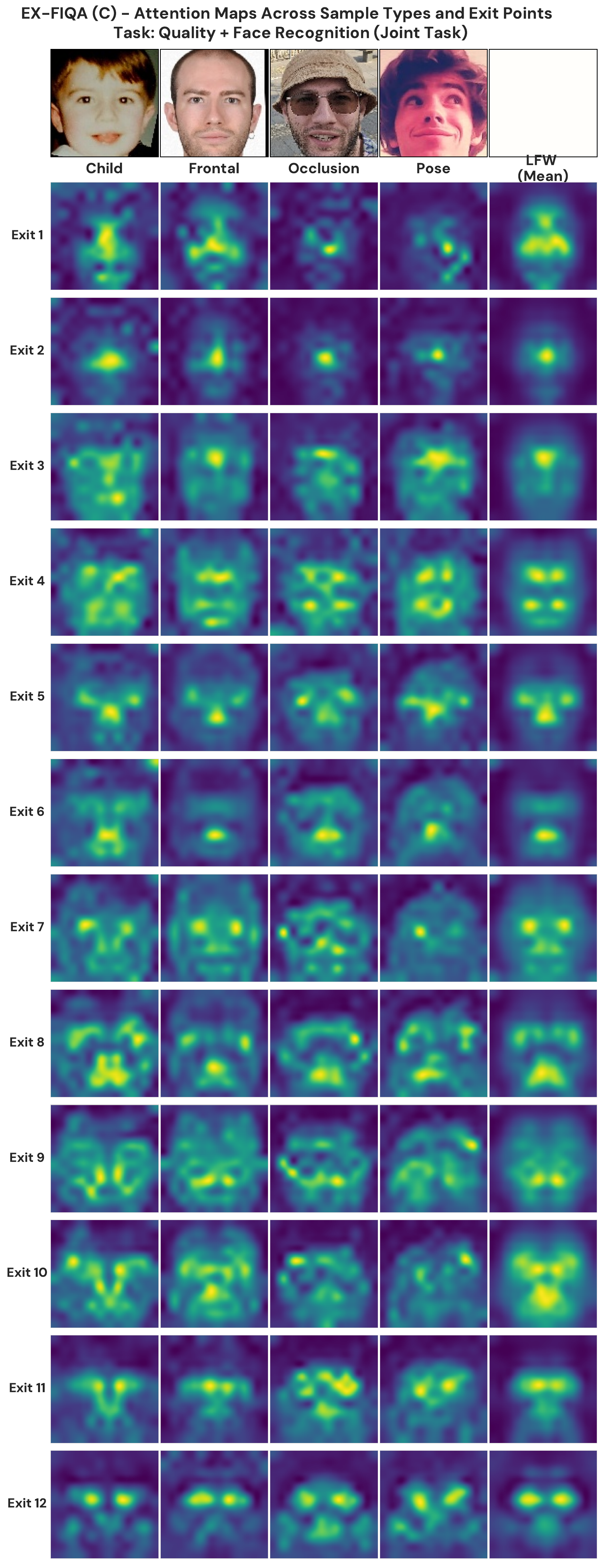}
    \caption{\earlyexit (C) attention map evolution across sample types and exit points. Visualization of attention patterns for four representative face images: a child face, a frontal pose, an occluded face, and a challenging pose. The top row displays original input images, followed by attention maps from Exit 1 through Exit 12. \textit{The individual appearing in the images have provided informed consent for their use in this research.}}
    \label{fig:crfiqa_axis2_sample}
\end{figure}

\begin{figure}[h]
    \centering
    \includegraphics[width=0.4\textwidth]{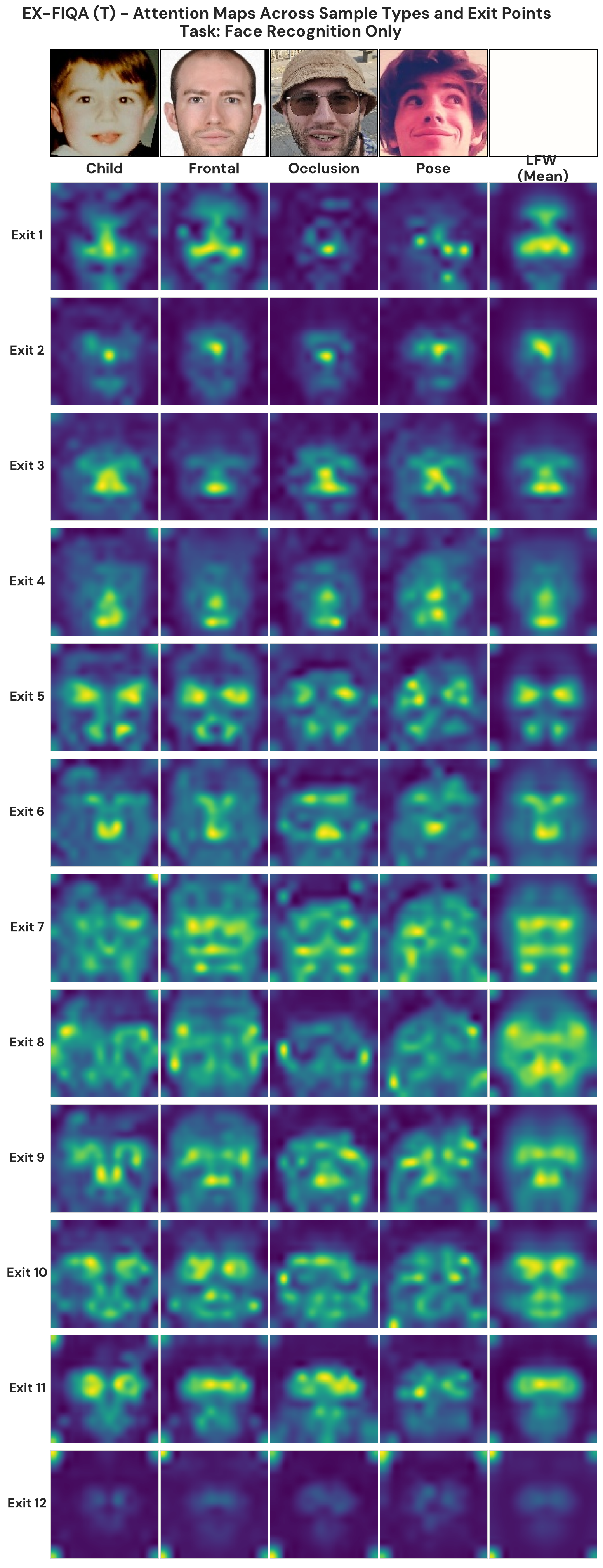}
    \caption{\earlyexit (T) attention map evolution across sample types and exit points. Visualization of attention patterns for four representative face images: a child face, a frontal pose, an occluded face, and a challenging pose. The top row displays original input images, followed by attention maps from Exit 1 through Exit 12. \textit{The individual appearing in the images have provided informed consent for their use in this research.}}
    \label{fig:token_axis2_sample}
\end{figure}

\begin{figure}[ht]
    \centering
    \includegraphics[width=0.5\textwidth]{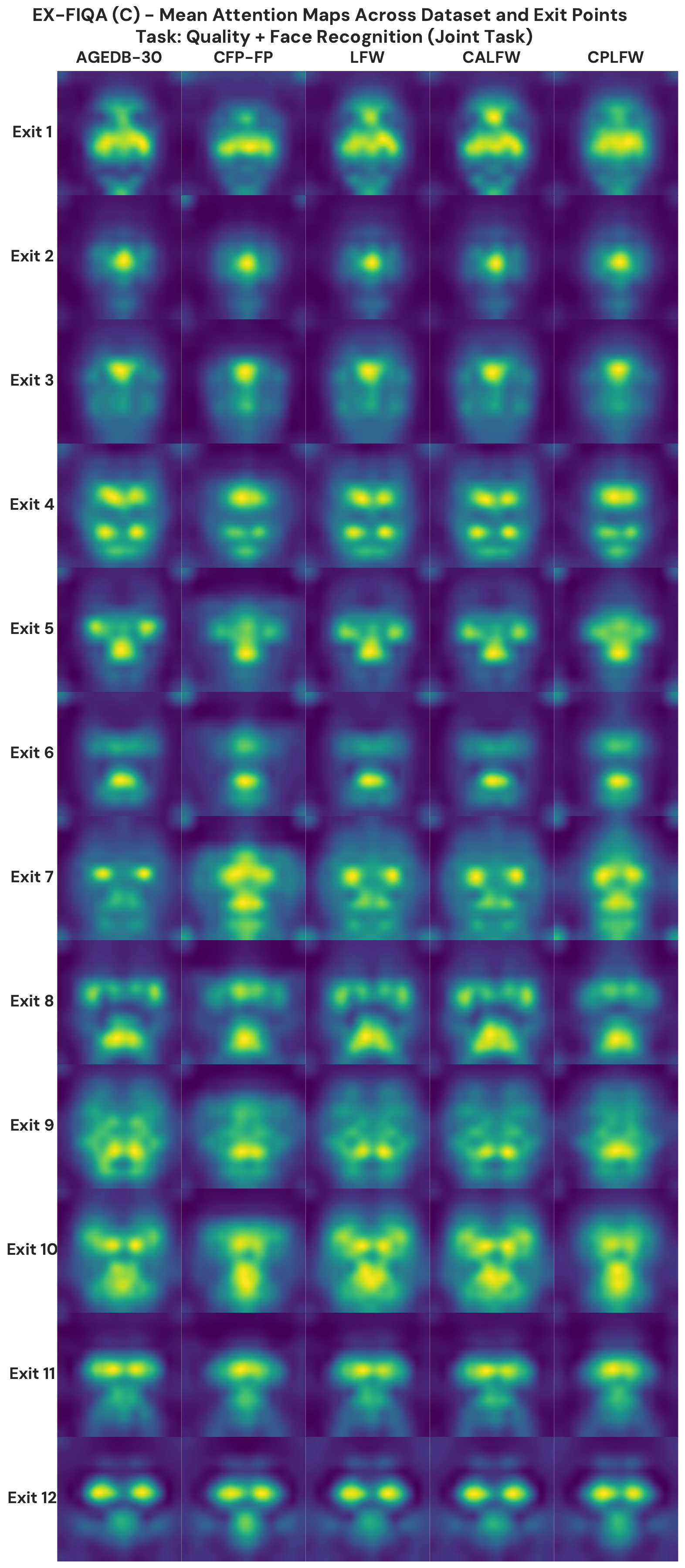}
    \caption{\earlyexit (C) model mean attention maps across datasets and exit points. Rows represent different exit points (1-12), while columns represent different datasets (AgeDB-30 \cite{agedb}, CALFW \cite{CALFW}, CFP-FP \cite{cfp-fp}, CPLFW \cite{CPLFWTech}, LFW \cite{LFWTech}).}
    \label{fig:crfiqa_axis2}
\end{figure}

\begin{figure}[ht]
    \centering
    \includegraphics[width=0.5\textwidth]{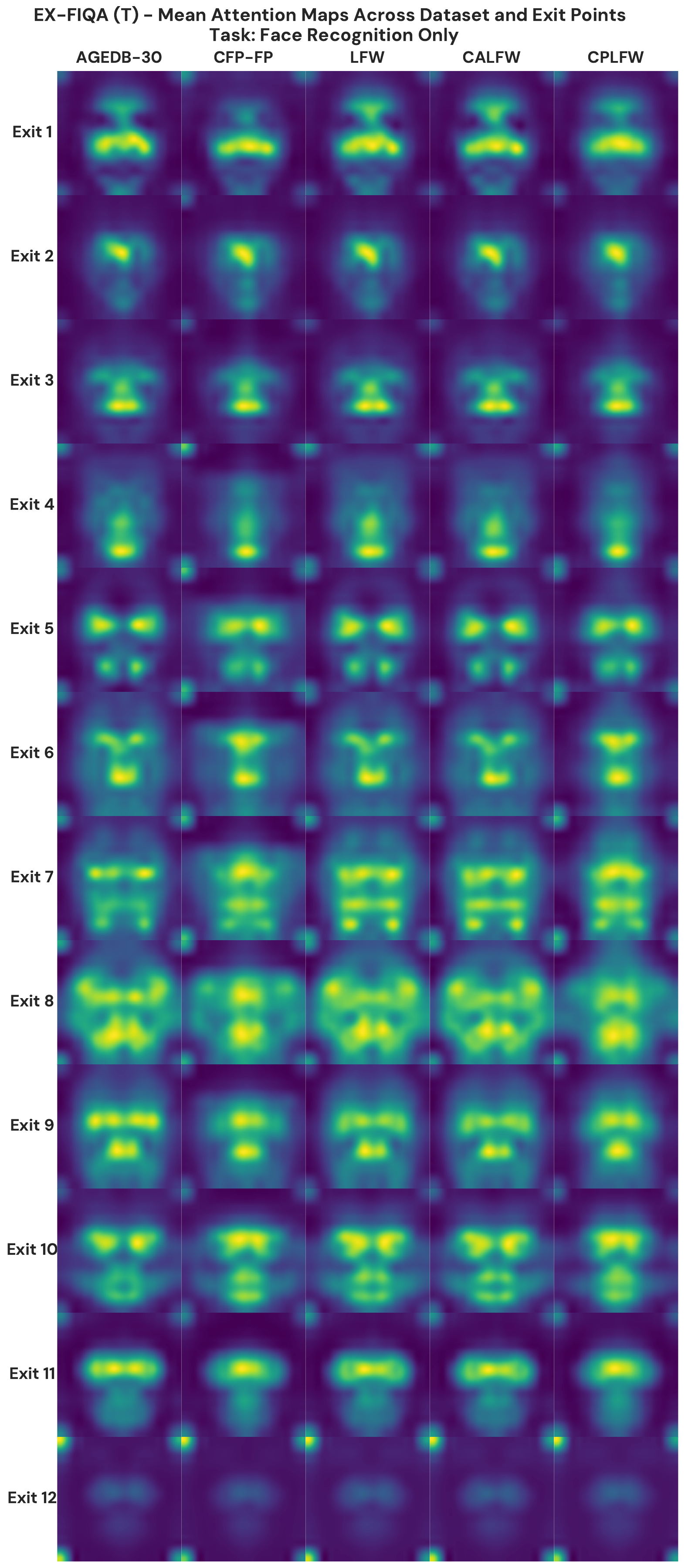}
    \caption{\earlyexit (T) model mean attention maps across datasets and exit points. Rows represent different exit points (1-12), while columns represent different datasets (AgeDB-30 \cite{agedb}, CALFW \cite{CALFW}, CFP-FP \cite{cfp-fp}, CPLFW \cite{CPLFWTech}, LFW \cite{LFWTech}).}
    \label{fig:token_axis2}
\end{figure}

\begin{figure}[htbp]
    \centering
    \includegraphics[width=0.5\textwidth]{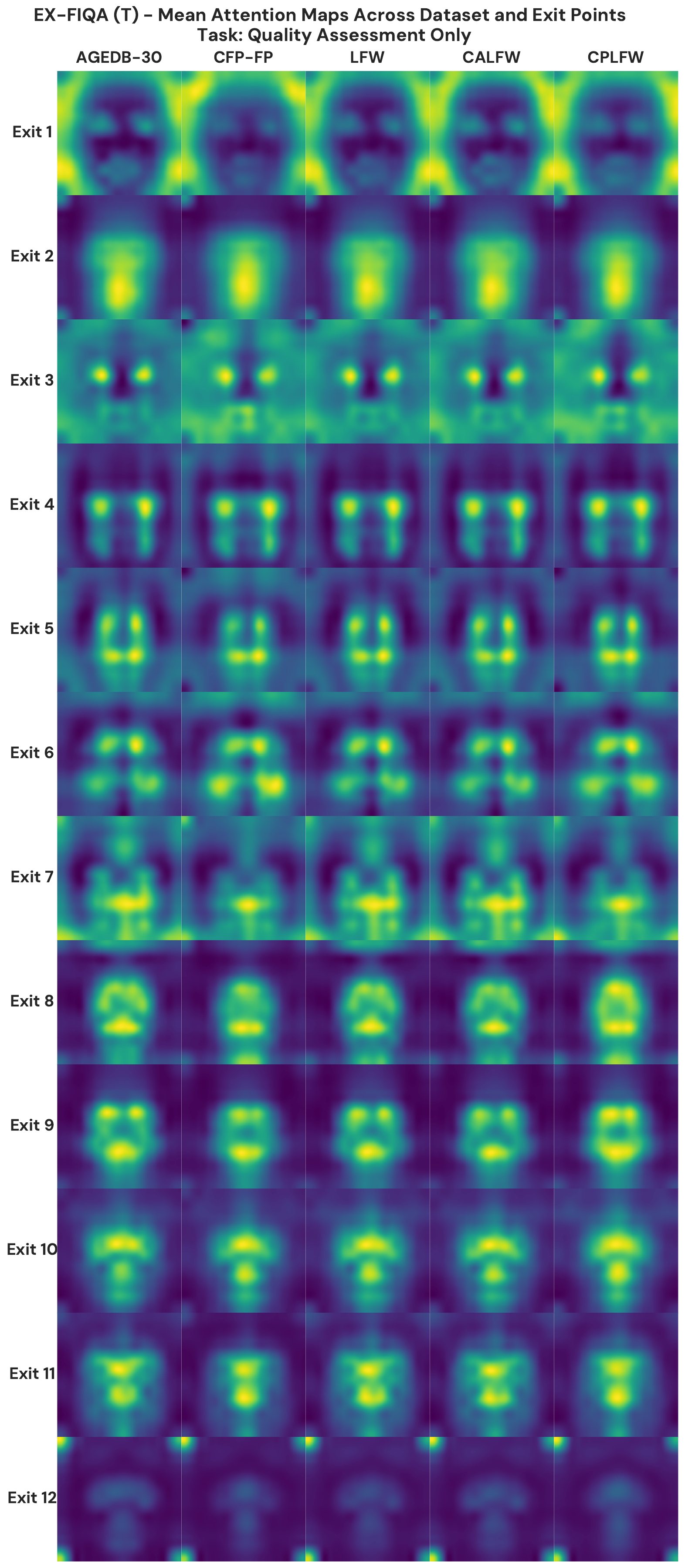}
    \caption{\earlyexit (T) model mean attention maps across datasets and exit points. Rows represent different exit points (1-12), while columns represent different datasets (AgeDB-30 \cite{agedb}, CALFW \cite{CALFW}, CFP-FP \cite{cfp-fp}, CPLFW \cite{CPLFWTech}, LFW \cite{LFWTech}).}
    \label{fig:token_axis4}
\end{figure}

\begin{figure*}[htbp]
    \centering
    \includegraphics[width=0.85\textwidth]{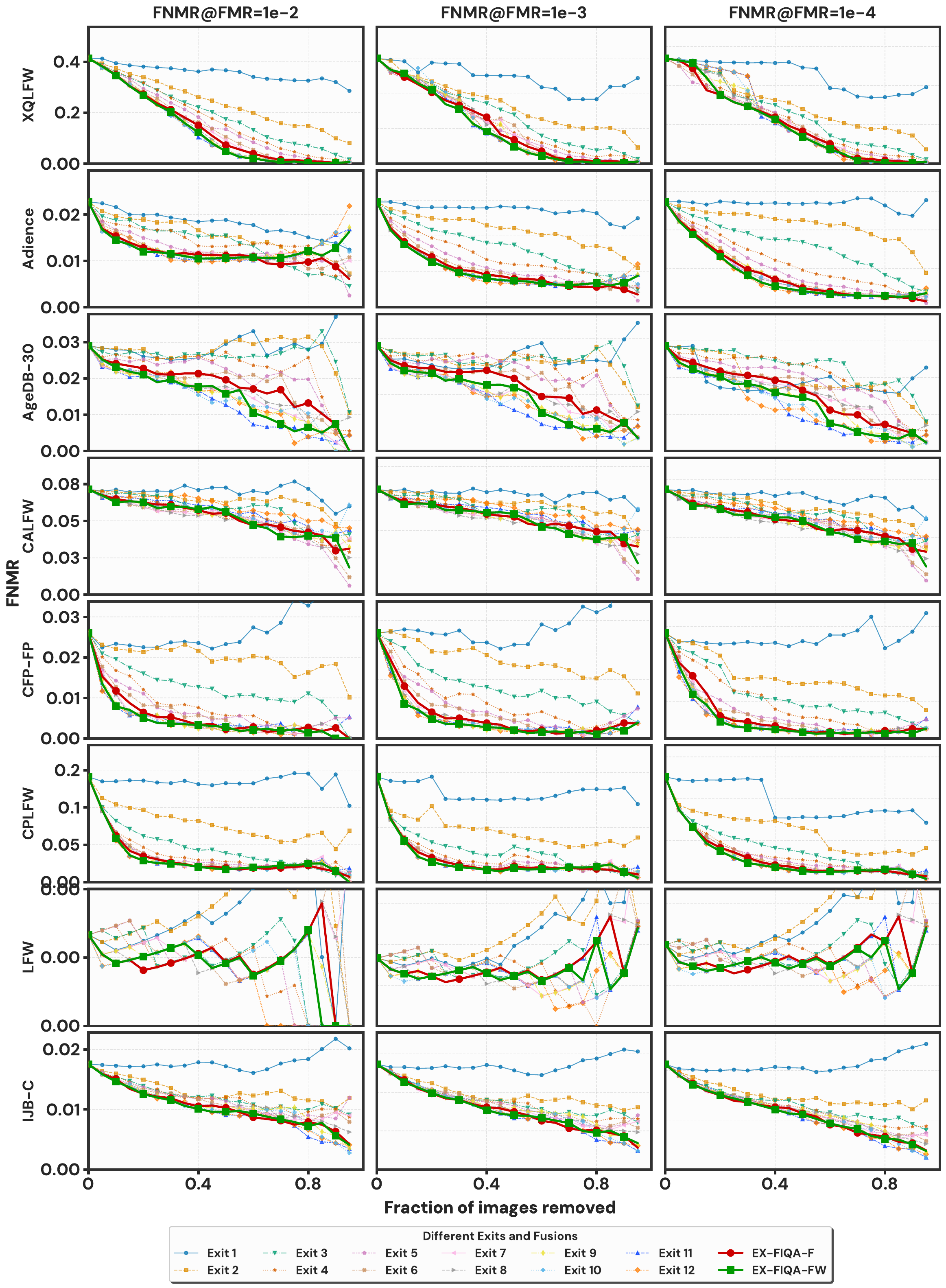}
    \caption{Error-versus-Discard Characteristic (EDC) curves for \earlyexit (C) ablation. The plot shows FNMR as a function of the fraction of images removed across eight FR datasets (LFW \cite{LFWTech}, AgeDB-30 \cite{agedb}, CFP-FP \cite{cfp-fp}, CALFW \cite{CALFW}, Adience \cite{Adience}, CPLFW \cite{CPLFWTech}, XQLFW \cite{XQLFW}, and IJB-C \cite{ijbc}) at three False Match Rate thresholds (FNMR@FMR=$1e-2$, $1e-3$, $1e-4$). Results are shown for \earlyexit (C), including 12 different exit points (Exit 1-12) and two fusion methods (\fusion: average fusion of all exits, \fusionw: weighted average fusion).}
    \label{fig:erc_curves_ablation_c}
\end{figure*}

\begin{figure*}[htbp]
    \centering
    \includegraphics[width=0.85\textwidth]{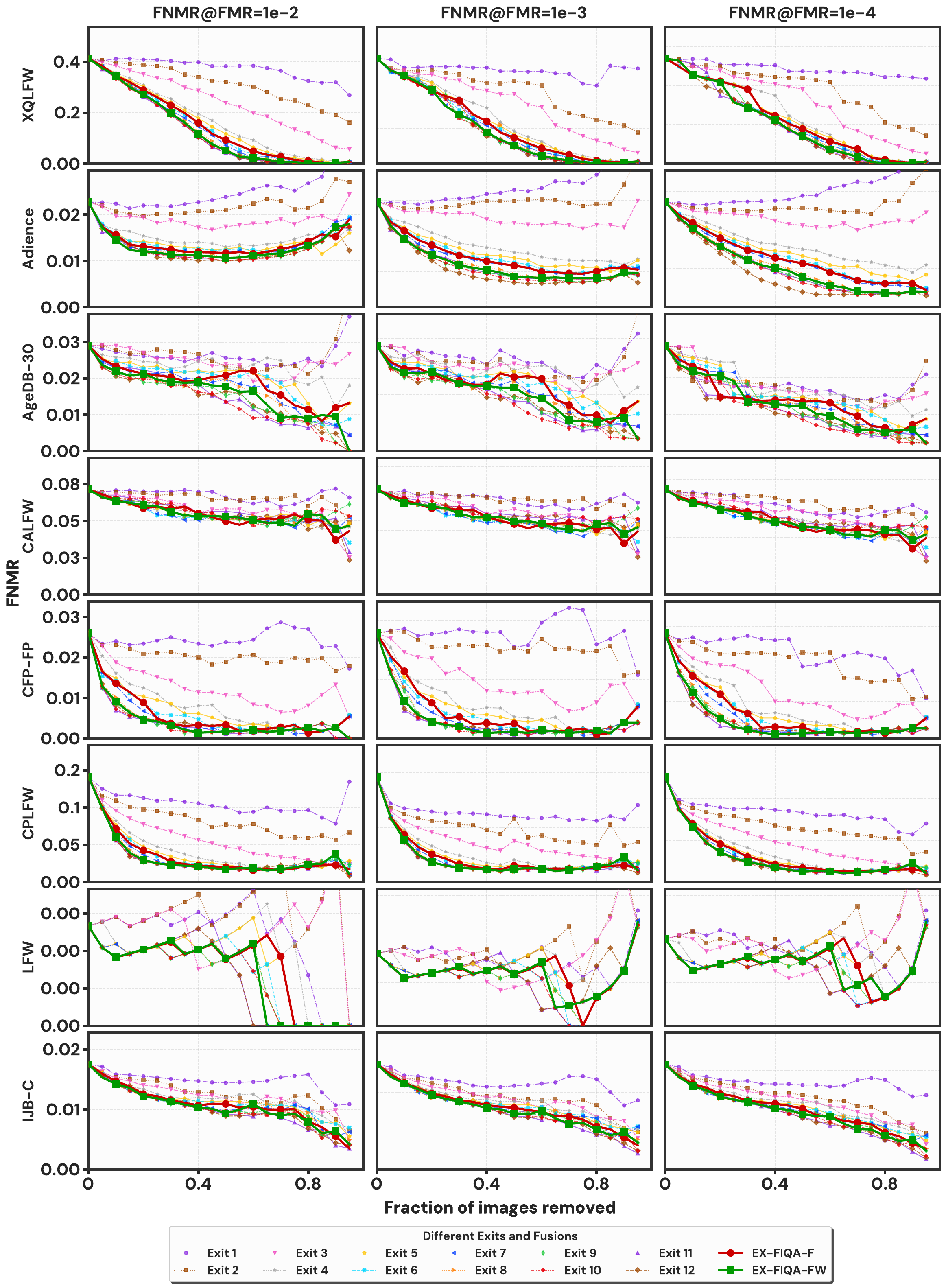}
    \caption{Error-versus-Discard Characteristic (EDC) curves for \earlyexit (T) ablation. The plot shows FNMR as a function of the fraction of images removed across eight FR datasets (LFW \cite{LFWTech}, AgeDB-30 \cite{agedb}, CFP-FP \cite{cfp-fp}, CALFW \cite{CALFW}, Adience \cite{Adience}, CPLFW \cite{CPLFWTech}, XQLFW \cite{XQLFW}, and IJB-C \cite{ijbc}) at three False Match Rate thresholds (FNMR@FMR=$1e-2$, $1e-3$, $1e-4$). Results are shown for \earlyexit (T), including 12 different exit points (Exit 1-12) and two fusion methods (\fusion: average fusion of all exits, \fusionw: weighted average fusion).}
    \label{fig:erc_curves_ablation_t}
\end{figure*}

\begin{figure*}[htbp]
    \centering
    \includegraphics[width=0.85\textwidth]{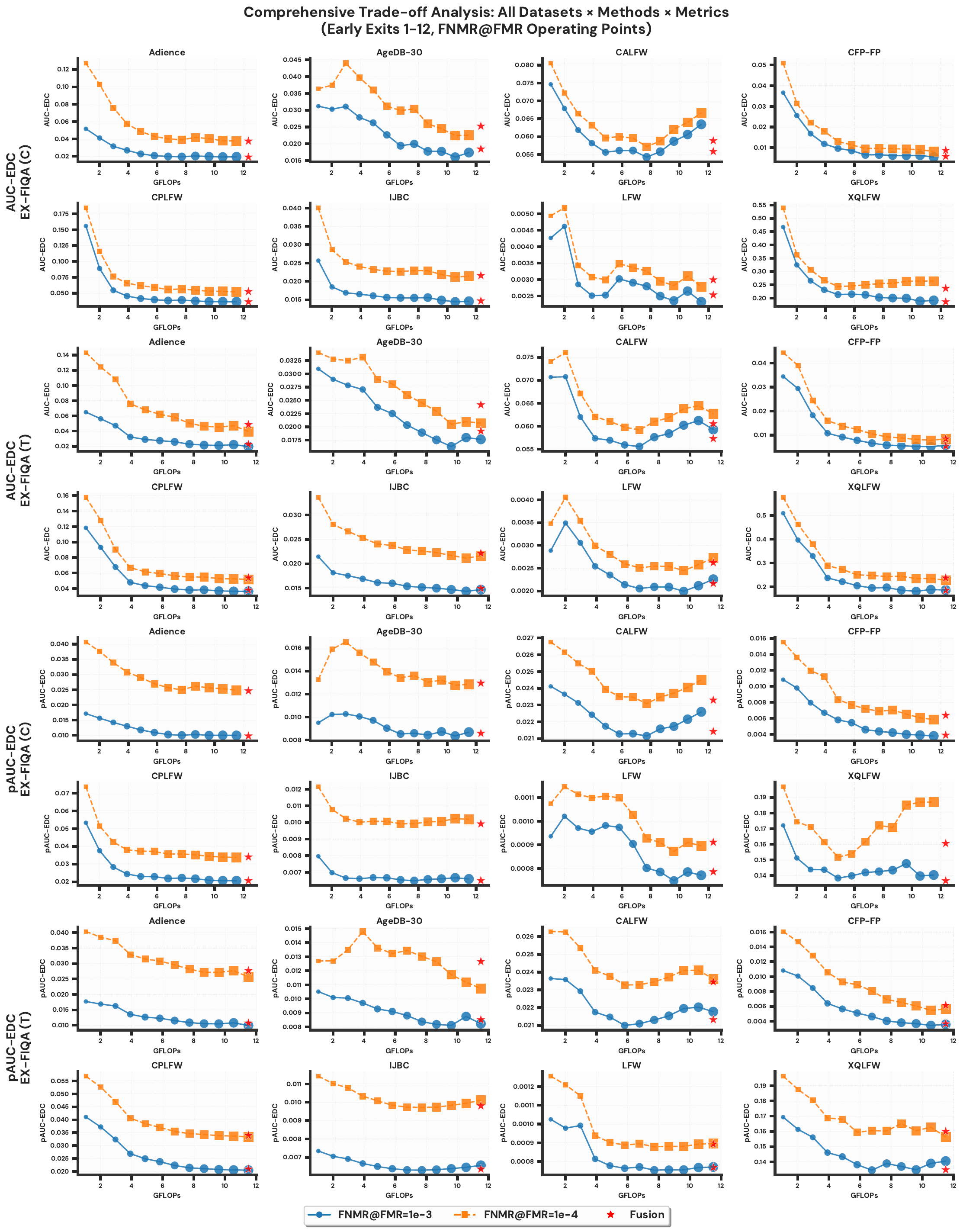}
    \caption{pAUC-EDC and AUC-EDC, FIQA performance metrics, for \earlyexit (C) and \earlyexit (T)  are shown for each exit across multiple datasets, including the computational cost. The trade-off between computational complexity (GFLOPs) and pAUC-EDC/AUC-EDC (FNMR@FMR=$1e-3$, $1e-4$) is illustrated. Deeper exits generally improve pAUC-EDC/AUC-EDC (the lower the better), albeit at a higher computational cost.}
    \label{fig:comprehensive}
\end{figure*}

\begin{figure*}[t]
    \centering
    \setlength{\tabcolsep}{1pt}
    \renewcommand{\arraystretch}{1.1}
    \resizebox{0.95\textwidth}{!}{
        \includegraphics[width=0.44\textwidth]{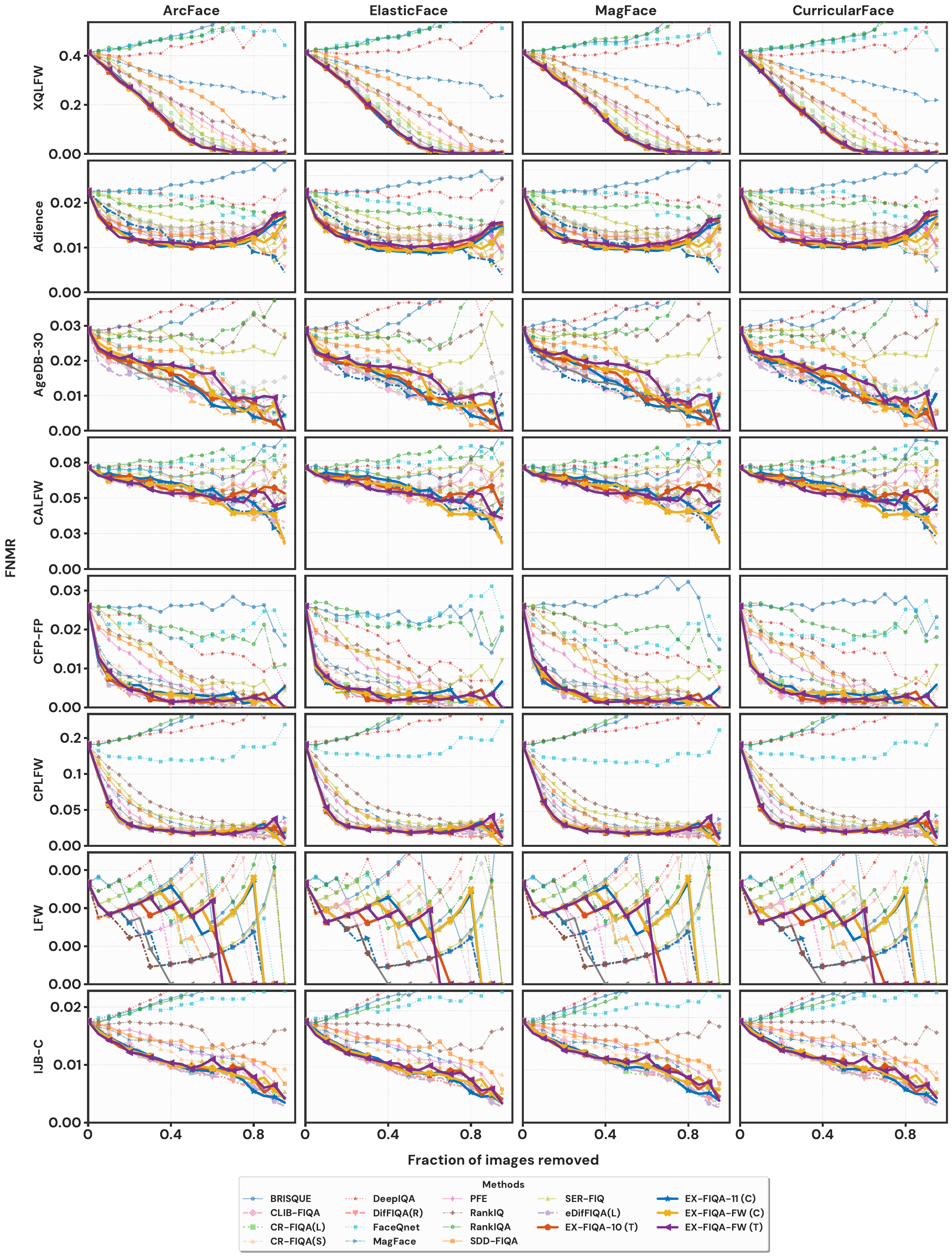}}
    \caption{Error-versus-Discard Characteristic (EDC) curves for FNMR@FMR=$1e-2$ of our proposed method in comparison to SOTA. Results shown on eight benchmark datasets: LFW \cite{LFWTech}, AgeDB-30 \cite{agedb}, CFP-FP \cite{cfp-fp}, CALFW \cite{CALFW}, Adience \cite{Adience}, CPLFW \cite{CPLFWTech}, XQLFW \cite{XQLFW}, and IJB-C \cite{ijbc}, using ArcFace \cite{deng2019arcface}, ElasticFace \cite{elasticface}, MagFace \cite{MagFace}, and CurricularFace \cite{curricularFace} FR models. The best exits, \earlyexit and proposed fusion \fusionw are marked with solid lines.}
    \label{fig:fnmr2}
\end{figure*}

\begin{figure*}[t]
    \centering
    \setlength{\tabcolsep}{1pt}
    \renewcommand{\arraystretch}{1.1}
    \resizebox{0.95\textwidth}{!}{
        \includegraphics[width=0.44\textwidth]{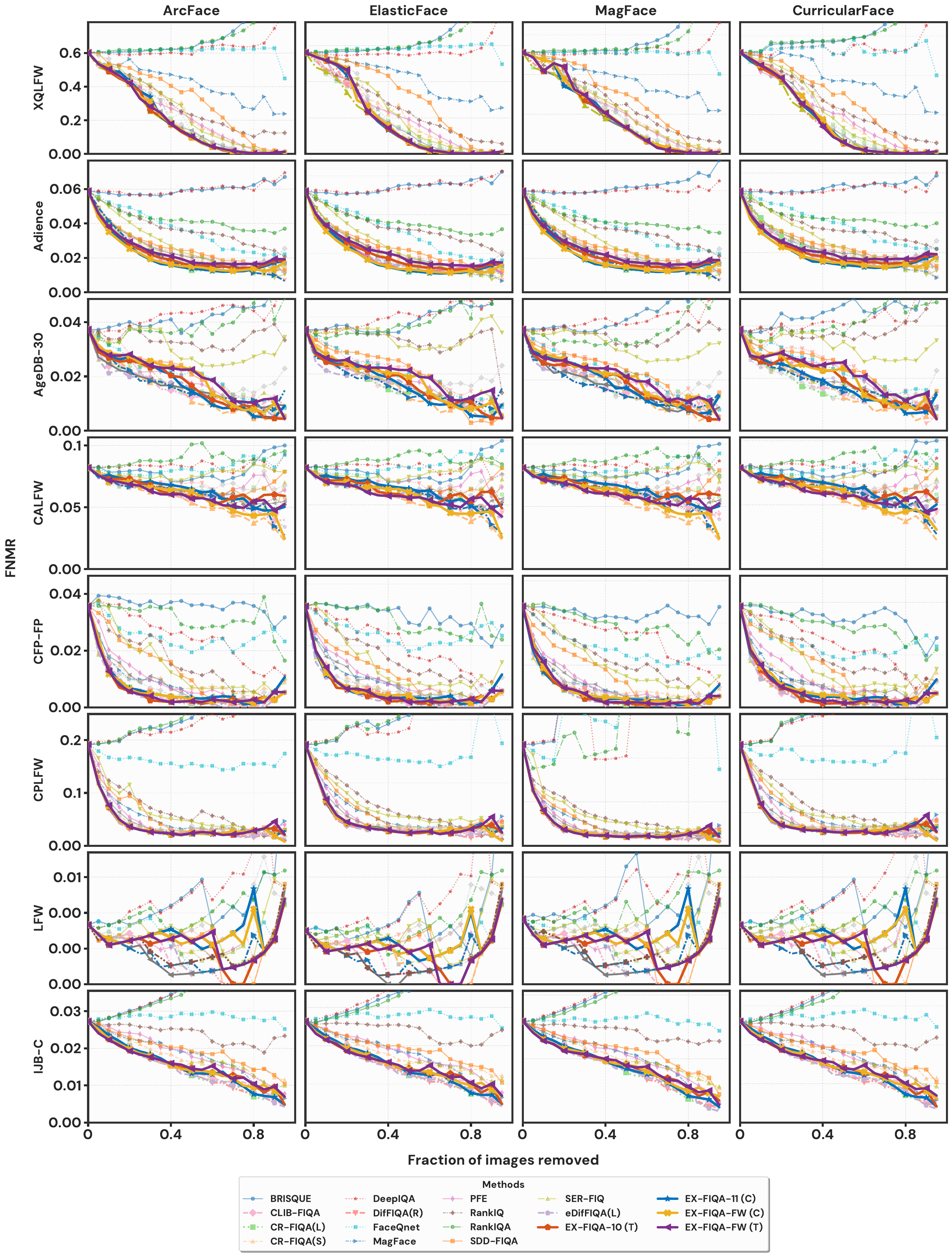}}
    \caption{Error-versus-Discard Characteristic (EDC) curves for FNMR@FMR=$1e-3$ of our proposed method in comparison to SOTA. Results shown on eight benchmark datasets: LFW \cite{LFWTech}, AgeDB-30 \cite{agedb}, CFP-FP \cite{cfp-fp}, CALFW \cite{CALFW}, Adience \cite{Adience}, CPLFW \cite{CPLFWTech}, XQLFW \cite{XQLFW}, and IJB-C \cite{ijbc}, using ArcFace \cite{deng2019arcface}, ElasticFace \cite{elasticface}, MagFace \cite{MagFace}, and CurricularFace \cite{curricularFace} FR models. The best exits, \earlyexit and proposed fusion \fusionw are marked with solid lines.}
    \label{fig:fnmr3}
\end{figure*}

\begin{figure*}[t]
    \centering
    \setlength{\tabcolsep}{1pt}
    \renewcommand{\arraystretch}{1.1}
    \resizebox{0.95\textwidth}{!}{
        \includegraphics[width=0.44\textwidth]{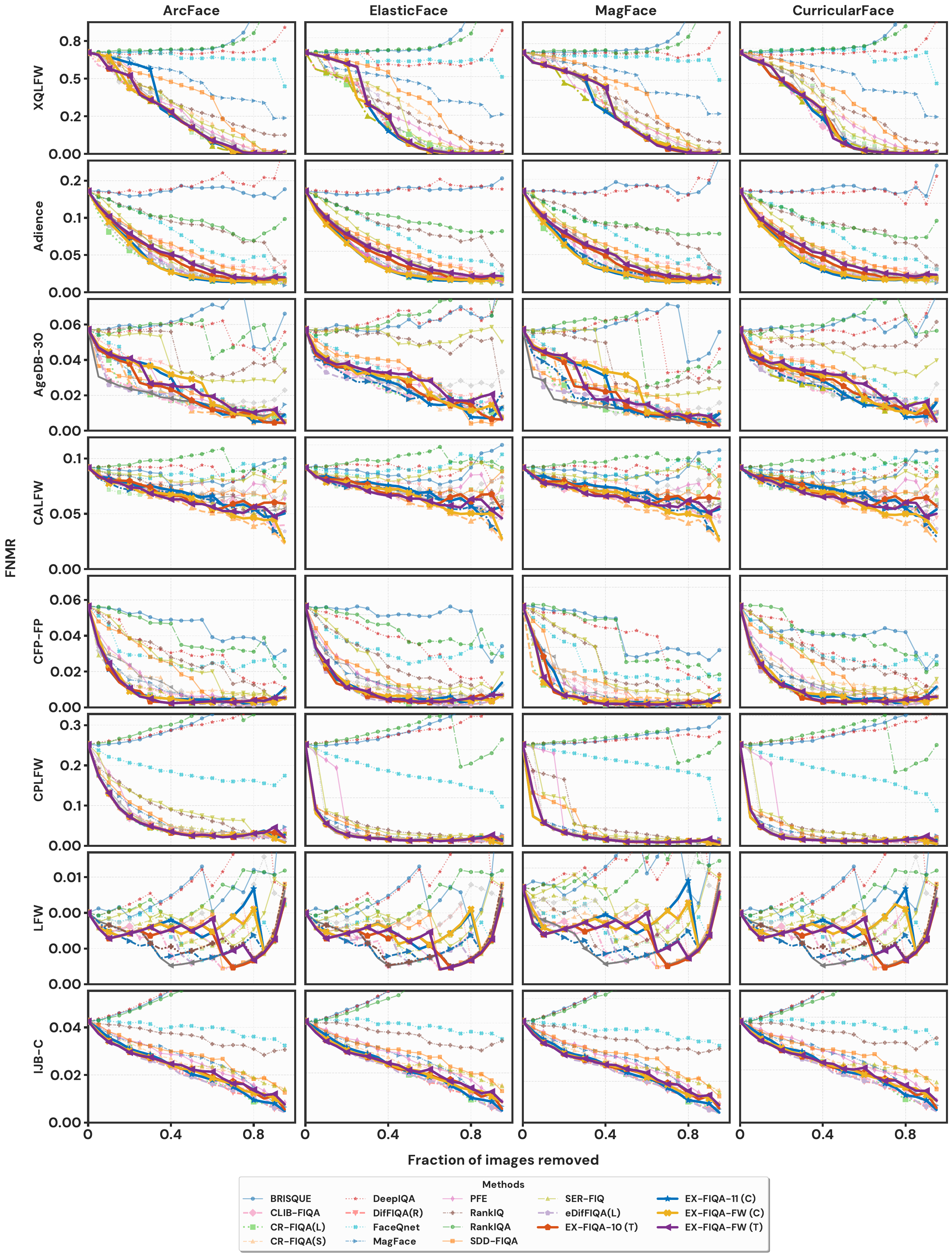}}
    \caption{Error-versus-Discard Characteristic (EDC) curves for FNMR@FMR=$1e-4$ of our proposed method in comparison to SOTA. Results shown on eight benchmark datasets: LFW \cite{LFWTech}, AgeDB-30 \cite{agedb}, CFP-FP \cite{cfp-fp}, CALFW \cite{CALFW}, Adience \cite{Adience}, CPLFW \cite{CPLFWTech}, XQLFW \cite{XQLFW}, and IJB-C \cite{ijbc}, using ArcFace \cite{deng2019arcface}, ElasticFace \cite{elasticface}, MagFace \cite{MagFace}, and CurricularFace \cite{curricularFace} FR models. The best exits, \earlyexit and proposed fusion \fusionw are marked with solid lines.}
    \label{fig:fnmr4}
\end{figure*}

\textbf{Figures: Visual Evidence and Insights}
\begin{itemize}
    \item \textbf{Figure \ref{fig:crfiqa_axis2_sample}}: Attention map evolution for \earlyexit (C) showing how attention patterns develop across 12 exits for four representative face types (child, frontal, occluded, challenging pose). \textbf{Figure \ref{fig:token_axis2_sample}}: Attention map evolution for \earlyexit (T) demonstrating attention pattern progression across exits for the same representative face samples. These attention visualizations are crucial to understand the internal mechanisms of our approach, providing interpretable evidence of how the model progressively focuses on different facial regions.

    \item \textbf{Figure \ref{fig:crfiqa_axis2}}: Dataset-level mean attention maps for \earlyexit (C) across five benchmark datasets (AgeDB-30, CALFW, CFP-FP, CPLFW, LFW) and all 12 exit points. \textbf{Figure \ref{fig:token_axis2}}: Dataset-level mean attention maps for \earlyexit (T) showing attention patterns across the same datasets and exit points. \textbf{Figure \ref{fig:token_axis4}}: Additional attention visualization for \earlyexit (T) quality assessment task across datasets and exit points. These comprehensive visualizations serve to provide understanding of the model across different data distributions, proving that our model learns consistent, dataset-agnostic attention patterns that focus on biometrically relevant facial regions rather than dataset-specific artifacts.

    \item \textbf{Figure \ref{fig:erc_curves_ablation_c}}: Error-versus-Discard Characteristic (EDC) curves for \earlyexit (C) ablation study showing FNMR performance across different discard fractions for all exits and fusion methods. \textbf{Figure \ref{fig:erc_curves_ablation_t}}: EDC curves for \earlyexit (T) ablation study with similar analysis structure as the C variant. These comprehensive EDC curves are fundamental to validate our core contribution, the effectiveness of early exits for quality assessment, showing readers exactly how performance varies across the transformer depth and justifying our architectural choices.

    \item \textbf{Figure \ref{fig:comprehensive}}: Comprehensive performance-complexity trade-off analysis showing pAUC-EDC and AUC-EDC metrics versus computational cost (GFLOPs) for both model variants across multiple datasets. This analysis addresses a critical practical concern, computational efficiency, enabling practitioners to balance quality assessment accuracy against computational constraints for informed deployment decisions in resource-limited scenarios.

    \item \textbf{Figures \ref{fig:fnmr2}, \ref{fig:fnmr3}, \ref{fig:fnmr4}}: State-of-the-art comparison EDC curves at three different FNMR@FMR thresholds ($1e-2$, $1e-3$, $1e-4$) across eight benchmark datasets and four face recognition models, comparing our method against existing approaches. These comprehensive comparisons provide definitive performance validation at multiple operating points, demonstrating that our competitiveness holds across various security requirements.
\end{itemize}

\input{tab/quality_scores}

\input{tab/ablation_auc}
\input{tab/AUC_all}
\input{tab/ijbc1v1}

\end{document}

%% file: tab/ablation_pauc.tex
\begin{table*}[ht!]
\begin{center}
\caption{Performance analysis of \earlyexit exits and fusion strategies. pAUC-EDC values are reported for individual exits (Exit 1-12) and fusion approaches (\fusion, \fusionw) across eight benchmark datasets using ArcFace as the face recognition model. Lower pAUC values indicate better quality assessment performance. Notations $1e-3$ and $1e-4$ indicate the fixed FMR thresholds. All values are scaled by a $10^3$ factor for better readability.}
\label{tab:ablation_pauc}
\resizebox{\textwidth}{!}{
\begin{tabular}{|l|ll|c|c|c|cc|cc|cc|cc|cc|cc|cc|cc|cc|}
\hline
\multirow{2}{*}{FR} &\multicolumn{2}{c|}{\multirow{2}{*}{Method}}  & \multirow{2}{*}{GFLOPs} & Ratio of&  \multirow{2}{*}{Params (M)} &\multicolumn{2}{c|}{Adience \cite{Adience}} & \multicolumn{2}{c|}{AgeDB-30 \cite{agedb}} & \multicolumn{2}{c|}{CFP-FP \cite{cfp-fp}} & \multicolumn{2}{c|}{LFW \cite{LFWTech}} & \multicolumn{2}{c|}{CALFW \cite{CALFW}} & \multicolumn{2}{c|}{CPLFW \cite{CPLFWTech}} & \multicolumn{2}{c|}{XQLFW \cite{XQLFW}} & \multicolumn{2}{c|}{IJB-C \cite{ijbc}} & \multicolumn{2}{c|}{Mean} \\
 &  & & &GLOPs& & $1e{-3}$ & $1e{-4}$ & $1e{-3}$ & $1e{-4}$ & $1e{-3}$ & $1e{-4}$ & $1e{-3}$ & $1e{-4}$ & $1e{-3}$ & $1e{-4}$ & $1e{-3}$ & $1e{-4}$ & $1e{-3}$ & $1e{-4}$ & $1e{-3}$ & $1e{-4}$ & $1e{-3}$ & $1e{-4}$ \\
\hline
\multirow{14}{*}{\rotatebox[origin=c]{90}{ArcFace\cite{deng2019arcface}}}  & \multirow{12}{*}{\rotatebox[origin=c]{90}{\earlyexit (C)}} 
&Exit 1	& 1.06	&	0.09 & 41.37 &             17.119	&	40.675	&	9.505	&	13.260	&	10.842	&	15.561	&	0.936	&	1.075	&	24.113	&	26.760	&	53.270	&	73.607	&	172.074	&	196.810	&	7.966	&	12.149	&	36.978	&	47.487	\\
&&Exit 2&2.01	&	0.17 &	44.52&               15.629	&	37.589	&	10.226	&	15.902	&	9.802	&	13.662	&	1.021	&	1.146	&	23.647	&	26.173	&	37.531	&	51.460	&	151.214	&	174.392	&	6.977	&	10.784	&	32.006	&	41.389	\\
&&Exit 3&2.96	&	0.26&	47.67&                14.227	&	33.936	&	10.273	&	16.522	&	7.956	&	11.977	&	0.971	&	1.114	&	23.130	&	25.495	&	28.312	&	42.446	&	143.832	&	171.110	&	6.656	&	10.234	&	29.419	&	39.104	\\
&&Exit 4&3.91	&	0.34&50.82	&            12.988	&	30.787	&	10.053	&	15.584	&	6.704	&	11.235	&	0.956	&	1.099	&	22.421	&	25.016	&	24.375	&	38.012	&	143.715	&	161.409	&	6.616	&	10.020	&	28.478	&	36.645	\\
&&Exit 5&4.85	&	0.42&	53.97&            11.778	&	28.989	&	9.714	&	14.801	&	5.803	&	8.352	&	0.982	&	1.107	&	21.736	&	23.938	&	22.982	&	37.288	&	138.360	&	151.748	&	6.685	&	10.069	&	27.255	&	34.536	\\
&&Exit 6	&5.80	&	0.50& 57.12   &            10.905	&	26.956	&	9.036	&	13.956	&	5.450	&	7.688	&	0.974	&	1.099	&	21.274	&	23.504	&	22.889	&	37.131	&	139.787	&	153.874	&	6.670	&	10.064	&	27.123	&	34.284 \\
&&Exit 7&6.75	&	0.59&	60.27&                10.244	&	25.658	&	8.531	&	13.407	&	4.585	&	7.188	&	0.904	&	1.028	&	21.290	&	23.468	&	21.862	&	35.608	&	141.945	&	161.617	&	6.544	&	9.921	&	26.988	&	34.737	\\
&&Exit 8&7.70	&	0.67&	63.42&            10.015	&	24.927	&	8.596	&	13.616	&	4.357	&	6.887	&	0.802	&	0.927	&	\textbf{21.152}	&	\textbf{23.087}	&	22.163	&	35.689	&	142.533	&	172.071	&	\textbf{6.493}	&	9.924	&	27.014	&	35.891	\\
&&Exit 9&8.65	&	0.75&	66.57&        10.284	&	26.169	&	8.438	&	13.029	&	4.207	&	7.037	&	0.785	&	0.910	&	21.579	&	23.477	&	21.712	&	35.178	&	143.430	&	170.729	&	6.575	&	10.039	&	27.126	&	35.821	\\
&&Exit 10&9.60	&	0.83&69.72	&          9.995	&	25.578	&	8.737	&	13.211	&	3.980	&	6.530	&	\textbf{0.748}	&	\textbf{0.873}	&	21.725	&	23.706	&	20.837	&	34.352	&	147.638	&	185.112	&	6.608	&	10.060	&	27.533	&	37.428	\\
&&Exit 11&10.55	&	0.92&	72.87&     10.023	&	25.330	&	\textbf{8.355}	&	\textbf{12.762}	&	3.902	&	6.055	&	0.785	&	0.910	&	22.154	&	24.054	&	20.637	&	33.948	&	139.682	&	186.923	&	6.670	&	10.240	&	26.526	&	37.528	\\
&&Exit 12&	11.49	&	1.00&   76.03&           9.948	&	24.833	&	8.692	&	12.853	&	\textbf{3.782}	&	\textbf{5.844}	&	0.771	&	0.896	&	22.594	&	24.504	&	\textbf{20.586}	&	\textbf{33.693}	&	140.344	&	187.017	&	6.599	&	10.185	&	26.664	&	37.478	\\ \cline{2-24} 
&\multicolumn{2}{c|}{\fusion (C)}&12.33	&	1.07&	76.03&    10.263	&	25.709	&	9.105	&	13.952	&	4.680	&	7.648	&	0.773	&	0.898	&	21.739	&	23.682	&	21.981	&	35.699	&	\textbf{136.317}	&	\textbf{154.694}	&	6.503	&	\textbf{9.867}	&	26.420	&	\textbf{34.019}	\\
&\multicolumn{2}{c|}{\fusionw (C)}	&12.33	&	1.07&76.03   &   \textbf{9.768}	&	\textbf{24.601}	&	8.586	&	12.943	&	3.887	&	6.370	&	0.786	&	0.911	&	21.423	&	23.293	&	20.619	&	33.971	&	136.670	&	160.505	&	6.503	&	9.915	&	\textbf{26.030}	&	34.064	\\

\hline \hline
\multirow{14}{*}{\rotatebox[origin=c]{90}{ArcFace\cite{deng2019arcface}}}          & \multirow{12}{*}{\rotatebox[origin=c]{90}{\earlyexit (T)}} 
&Exit 1	&0.99	&	0.09& 41.37& 17.707	&	40.391	&	10.516	&	12.695	&	10.822	&	16.086	&	1.025	&	1.255	&	23.641	&	26.290	&	41.072	&	56.822	&	169.366	&	196.269	&	7.344	&	11.421	&	35.187	&	45.154	\\
&&Exit 2&1.95	&	0.17&44.52&	16.878	&	38.527	&	10.098	&	12.701	&	10.077	&	14.703	&	0.978	&	1.209	&	23.581	&	26.267	&	37.209	&	52.612	&	161.329	&	187.615	&	7.054	&	11.019	&	33.400	&	43.082	\\
&&Exit 3&2.90	&	0.25&47.67&	16.253	&	37.403	&	10.054	&	13.494	&	8.469	&	12.840	&	0.992	&	1.151	&	22.926	&	25.359	&	32.378	&	46.985	&	156.128	&	180.488	&	6.911	&	10.783	&	31.764	&	41.063\\
&&Exit 4&3.86	&	0.34&50.82&	13.552	&	32.922	&	9.707	&	14.795	&	6.402	&	10.586	&	0.814	&	0.938	&	21.747	&	24.098	&	26.892	&	40.624	&	146.041	&	168.770	&	6.664	&	10.332	&	28.977	&	37.883\\
&&Exit 5&4.81	&	0.42&53.97&	12.641	&	31.479	&	9.283	&	13.628	&	5.650	&	9.290	&	0.778	&	0.902	&	21.471	&	23.773	&	24.969	&	38.523	&	143.378	&	167.997	&	6.499	&	10.079	&	28.084	&	36.959	\\
&&Exit 6&5.77	&	0.50&57.12&	12.285	&	30.799	&	9.103	&	13.226	&	5.100	&	8.904	&	0.764	&	0.888	&	\textbf{20.998}	&	\textbf{23.280}	&	23.801	&	37.033	&	138.132	&	159.447	&	6.377	&	9.835	&	27.070	&	35.426	\\
&&Exit 7&6.72	&	0.58&60.27&	11.522	&	29.577	&	8.812	&	13.443	&	4.624	&	8.064	&	0.771	&	0.895	&	21.099	&	23.288	&	22.358	&	35.484	&	\textbf{134.511}	&	160.494	&	6.307	&	9.735	&	26.251	&	35.123\\
&&Exit 8&7.68	&	0.67&63.42	&10.868	&	28.188	&	8.377	&	13.020	&	4.035	&	6.936	&	\textbf{0.754}	&	\textbf{0.879}	&	21.306	&	23.454	&	21.427	&	34.642	&	138.908	&	160.310	&	\textbf{6.298}	&	\textbf{9.719}	&	26.496	&	34.644	\\
&&Exit 9&8.63	&	0.75&66.57&	10.531	&	27.167	&	8.187	&	12.648	&	3.804	&	6.529	&	0.756	&	0.881	&	21.541	&	23.726	&	21.120	&	34.292	&	136.914	&	165.112	&	6.322	&	9.737	&	26.147	&	35.011	\\
&&Exit 10&9.59	&	0.83&69.72&	10.459	&	27.127	&	\textbf{8.108}	&	11.713	&	3.672	&	6.055	&	0.756	&	0.881	&	21.952	&	24.095	&	20.782	&	33.871	&	134.815	&	160.456	&	6.384	&	9.830	&	\textbf{25.866}	&	34.253\\
&&Exit 11&10.54	&	0.92&72.88&	10.792	&	27.695	&	8.748	&	11.192	&	\textbf{3.426}	&	\textbf{5.453}	&	0.768	&	0.893	&	22.029	&	24.104	&	20.627	&	33.696	&	138.973	&	162.741	&	6.454	&	9.943	&	26.477	&	34.465	\\
&&Exit 12&	11.50	&	1.00&76.03&\textbf{9.948}	&	\textbf{25.664}	&	8.234	&	\textbf{10.734}	&	3.568	&	5.663	&	0.771	&	0.896	&	21.771	&	23.614	&	\textbf{20.531}	&	\textbf{33.388}	&	140.465	&	\textbf{156.275}	&	6.563	&	10.118	&	26.481	&	\textbf{33.294}	\\ \cline{2-24} 
& \multicolumn{2}{c|}{\fusion (T)}&11.50	&	1.00&76.03&	11.942	&	30.214	&	8.692	&	12.060	&	5.335	&	8.938	&	0.765	&	0.889	&	21.533	&	23.807	&	23.752	&	37.469	&	138.803	&	167.059	&	6.480	&	10.052	&	27.163	&	36.311	\\
& \multicolumn{2}{c|}{\fusionw (T)}&11.50	&	1.00&76.03&	10.700	&	27.716	&	8.515	&	12.656	&	3.639	&	6.137	&	0.768	&	0.893	&	21.322	&	23.451	&	20.943	&	33.935	&	134.790	&	160.018	&	6.362	&	9.802	&	25.880	&	34.326\\

\hline
\end{tabular}} 
\end{center}
\end{table*}

%% file: tab/pAUC_all.tex
\begin{table*}[ht!]
    \begin{center}
    \caption{The pAUCs of EDC achieved by our method and the SOTA methods under different experimental settings. The notions of $1e-3$ and $1e-4$ indicate the value of the fixed FMR at which the EDC curves (FNMR vs.~reject) were calculated. \earlyexit: Best Exit, \fusionw: Fusion. All values are scaled by a $10^3$ factor for better readability.}
    \label{tab:sota_pauc}
    \resizebox{0.99\textwidth}{!}{
\begin{tabular}{|c|cl|cc|cc|cc|cc|cc|cc|cc|cc|}
    \hline
\multirow{3}{*}{FR} & \multicolumn{2}{c|}{\multirow{3}{*}{Method}} 
& \multicolumn{2}{c|}{Adience} & \multicolumn{2}{c|}{AgeDB-30} & \multicolumn{2}{c|}{CFP-FP} 
& \multicolumn{2}{c|}{LFW} & \multicolumn{2}{c|}{CALFW} & \multicolumn{2}{c|}{CPLFW} 
& \multicolumn{2}{c|}{XQLFW} & \multicolumn{2}{c|}{IJB-C} \\ 
 & & 
 & \multicolumn{2}{c|}{\cite{Adience}} & \multicolumn{2}{c|}{\cite{agedb}} & \multicolumn{2}{c|}{\cite{cfp-fp}} 
 & \multicolumn{2}{c|}{\cite{LFWTech}} & \multicolumn{2}{c|}{\cite{CALFW}} & \multicolumn{2}{c|}{\cite{CPLFWTech}} 
 & \multicolumn{2}{c|}{\cite{XQLFW}}  & \multicolumn{2}{c|}{\cite{ijbc}} \\ \cline{4-19}
                                                                                    &                                                   &  & $1e{-3}$ & $1e{-4}$ & $1e{-3}$ & $1e{-4}$  & $1e{-3}$ & $1e{-4}$  & $1e{-3}$ & $1e{-4}$  & $1e{-3}$ & $1e{-4}$ & $1e{-3}$ & $1e{-4}$ & $1e{-3}$ & $1e{-4}$ & $1e{-3}$ & $1e{-4}$ \\ \hline
     \hline 
     \multirow{17}{*}{\rotatebox[origin=c]{90}{ArcFace\cite{deng2019arcface}}}      & \multirow{3}{*}{\rotatebox[origin=c]{90}{IQA}}    & \multicolumn{1}{|l|}{BRISQUE\cite{BRISQE_IQA}} &17.1746 &39.8035 &11.5267 &17.4919 &11.3476 &16.2225 &1.0438 &1.2273 &23.9653 &26.8772 &61.1376 &77.2797 &181.8562 &203.0627 &8.7528 &13.7855                                                \\
                                                                                    &                                                   & \multicolumn{1}{|l|}{RankIQA\cite{liu2017rankiqa}} &14.6239 &35.4490 &10.5402 &17.0561 &10.5550 &15.9389 &1.0739 &1.2560 &25.2499 &28.2475 &61.6504 &80.0431 &184.0819 &205.2308 &8.6993 &13.6493                                                \\
\hline 
    \end{tabular}}
    \end{center}
    \end{table*}

%% file: tab/quality_scores.tex
\begin{table}[htbp]
\begin{center}
\caption{Quality score evolution across transformer exits for representative face samples. Higher scores indicate better predicted quality. Rankings in parentheses show relative quality order for each exit (1=highest, 4=lowest).}
\label{tab:quality_scores}
\vspace{-2mm}
\resizebox{0.8\linewidth}{!}{
\begin{tabular}{|c|c|c|c|c|}
\hline
\multirow{2}{*}{Exit} & \multicolumn{4}{c|}{\earlyexit (C)} \\
\cline{2-5}
 & Image 1 (Child) & Image 2 (Frontal) & Image 3 (Occlusion) & Image 4 (Pose) \\
\hline
1 & 0.3418 (3) & 0.3949 (2) & 0.4199 (1) & 0.3135 (4) \\
2 & 0.2394 (3) & 0.3191 (1) & 0.3090 (2) & 0.2213 (4) \\
3 & 0.1945 (3) & 0.2543 (1) & 0.2352 (2) & 0.1875 (4) \\
4 & 0.1643 (3) & 0.2108 (1) & 0.2049 (2) & 0.1265 (4) \\
5 & 0.1792 (3) & 0.2206 (1) & 0.2054 (2) & 0.1400 (4) \\
6 & 0.1844 (3) & 0.2517 (1) & 0.2320 (2) & 0.1506 (4) \\
7 & 0.2025 (3) & 0.2813 (1) & 0.2625 (2) & 0.1987 (4) \\
8 & 0.2246 (4) & 0.3302 (1) & 0.2721 (2) & 0.2444 (3) \\
9 & 0.2760 (3) & 0.3926 (1) & 0.3147 (2) & 0.2560 (4) \\
10 & 0.3712 (2) & 0.5156 (1) & 0.3702 (3) & 0.3214 (4) \\
11 & 0.4905 (2) & 0.6159 (1) & 0.4363 (3) & 0.3915 (4) \\
12 & 0.5002 (2) & 0.6210 (1) & 0.4829 (3) & 0.4266 (4) \\
\hline \hline
\multirow{2}{*}{Exit} & \multicolumn{4}{c|}{\earlyexit (T)} \\
\cline{2-5}
 & Image 1 (Child) & Image 2 (Frontal) & Image 3 (Occlusion) & Image 4 (Pose) \\
\hline
1 & 0.4073 (2) & 0.4105 (1) & 0.4026 (3) & 0.3956 (4) \\
2 & 0.4056 (3) & 0.4129 (1) & 0.4103 (2) & 0.3903 (4) \\
3 & 0.4088 (3) & 0.4220 (1) & 0.4097 (2) & 0.3942 (4) \\
4 & 0.4089 (2) & 0.4314 (1) & 0.4071 (3) & 0.3918 (4) \\
5 & 0.4068 (2) & 0.4360 (1) & 0.4057 (3) & 0.3912 (4) \\
6 & 0.4096 (2) & 0.4441 (1) & 0.4047 (3) & 0.3882 (4) \\
7 & 0.4138 (2) & 0.4512 (1) & 0.4029 (3) & 0.3872 (4) \\
8 & 0.4049 (3) & 0.4510 (1) & 0.4090 (2) & 0.3782 (4) \\
9 & 0.4067 (4) & 0.4745 (1) & 0.4225 (2) & 0.3804 (3) \\
10 & 0.4277 (3) & 0.5148 (1) & 0.4303 (2) & 0.3966 (4) \\
11 & 0.4724 (2) & 0.5803 (1) & 0.4671 (3) & 0.4457 (4) \\
12 & 0.4847 (3) & 0.6043 (1) & 0.4889 (2) & 0.4785 (4) \\
\hline
\end{tabular}}
\vspace{-2mm}
\end{center}
\end{table}

%% file: tab/ablation_auc.tex
\begin{table*}[ht!]
\begin{center}
\caption{Performance analysis of \earlyexit exits and fusion strategies. AUC-EDC values are reported for individual exits (Exit 1-12) and fusion approaches (\fusion, \fusionw) across eight benchmark datasets using ArcFace as the face recognition model. Lower AUC values indicate better quality assessment performance. Notations $1e-3$ and $1e-4$ indicate the fixed FMR thresholds.}
\label{tab:ablation_auc}
\vspace{-2mm}
\resizebox{\textwidth}{!}{
\begin{tabular}{|l|ll|cc|cc|cc|cc|cc|cc|cc|cc|cc|}
\hline
\multirow{2}{*}{FR} &\multicolumn{2}{c|}{\multirow{2}{*}{Method}}  & \multicolumn{2}{c|}{Adience \cite{Adience}} & \multicolumn{2}{c|}{AgeDB-30 \cite{agedb}} & \multicolumn{2}{c|}{CFP-FP \cite{cfp-fp}} & \multicolumn{2}{c|}{LFW \cite{LFWTech}} & \multicolumn{2}{c|}{CALFW \cite{CALFW}} & \multicolumn{2}{c|}{CPLFW \cite{CPLFWTech}} & \multicolumn{2}{c|}{XQLFW \cite{XQLFW}} & \multicolumn{2}{c|}{IJB-C \cite{ijbc}} & \multicolumn{2}{c|}{Mean} \\
 &  & & $1e{-3}$ & $1e{-4}$ & $1e{-3}$ & $1e{-4}$ & $1e{-3}$ & $1e{-4}$ & $1e{-3}$ & $1e{-4}$ & $1e{-3}$ & $1e{-4}$ & $1e{-3}$ & $1e{-4}$ & $1e{-3}$ & $1e{-4}$ & $1e{-3}$ & $1e{-4}$ & $1e{-3}$ & $1e{-4}$ \\
\hline
\multirow{14}{*}{\rotatebox[origin=c]{90}{ArcFace\cite{deng2019arcface}}}  & \multirow{12}{*}{\rotatebox[origin=c]{90}{\earlyexit (C)}} & Exit 1    & 0.0517          & 0.1273          & 0.0312          & 0.0364          & 0.0366          & 0.0509          & 0.0043          & 0.0049          & 0.0746          & 0.0806          & 0.1559          & 0.1846          & 0.4672          & 0.5395          & 0.0257          & 0.0401          & 0.10590          & 0.13304          \\
             &  & Exit 2             & 0.0412          & 0.1028          & 0.0303          & 0.0375          & 0.0255          & 0.0314          & 0.0046          & 0.0052          & 0.0679          & 0.0722          & 0.0888          & 0.1161          & 0.3246          & 0.3622          & 0.0185          & 0.0287          & 0.07518          & 0.09451          \\
             &  & Exit 3                  & 0.0316          & 0.0761          & 0.0311          & 0.0440          & 0.0168          & 0.0222          & 0.0029          & 0.0034          & 0.0618          & 0.0664          & 0.0544          & 0.0762          & 0.2652          & 0.3068          & 0.0169          & 0.0253          & 0.06009          & 0.07755          \\
\hline
\end{tabular}}
\end{center}
\end{table*}

%% file: tab/AUC_all.tex
\begin{table*}[ht!]
    \begin{center}
    \caption{The AUCs of EDC achieved by our method and the SOTA methods under different experimental settings. The notions of $1e-3$ and $1e-4$ indicate the value of the fixed FMR at which the EDC curves (FNMR vs.~reject) were calculated. The results are compared to three IQA and twelve FIQA approaches. The XQLFW dataset uses SER-FIQ (marked with *) as the FIQ labeling method. \earlyexit: Best Exit, \fusionw: Fusion.}
    \label{tab:sota_auc}
    \vspace{-3mm}
    \resizebox{0.99\textwidth}{!}{
\begin{tabular}{|c|cl|cc|cc|cc|cc|cc|cc|cc|cc|}
    \hline
\multirow{3}{*}{FR} & \multicolumn{2}{c|}{\multirow{3}{*}{Method}} 
& \multicolumn{2}{c|}{Adience} & \multicolumn{2}{c|}{AgeDB-30} & \multicolumn{2}{c|}{CFP-FP} 
& \multicolumn{2}{c|}{LFW} & \multicolumn{2}{c|}{CALFW} & \multicolumn{2}{c|}{CPLFW} 
& \multicolumn{2}{c|}{XQLFW} & \multicolumn{2}{c|}{IJB-C}\\ 
 & & 
 & \multicolumn{2}{c|}{\cite{Adience}} & \multicolumn{2}{c|}{\cite{agedb}} & \multicolumn{2}{c|}{\cite{cfp-fp}} 
 & \multicolumn{2}{c|}{\cite{LFWTech}} & \multicolumn{2}{c|}{\cite{CALFW}} & \multicolumn{2}{c|}{\cite{CPLFWTech}} 
 & \multicolumn{2}{c|}{\cite{XQLFW}} & \multicolumn{2}{c|}{\cite{ijbc}} \\ \cline{4-19}
                                                                                    &                                                   &  & $1e{-3}$ & $1e{-4}$ & $1e{-3}$ & $1e{-4}$  & $1e{-3}$ & $1e{-4}$  & $1e{-3}$ & $1e{-4}$  & $1e{-3}$ & $1e{-4}$ & $1e^{-3}$ & $1e{-4}$ & $1e{-3}$ & $1e{-4}$ & $1e{-3}$ & $1e{-4}$\\ \hline
     \hline 
     \multirow{20}{*}{\rotatebox[origin=c]{90}{ArcFace\cite{deng2019arcface}}}      & \multirow{3}{*}{\rotatebox[origin=c]{90}{IQA}}    & \multicolumn{1}{|l|}{BRISQUE\cite{BRISQE_IQA}} &0.0565 & 0.1285 &0.0400 & 0.0585 &0.0343 & 0.0433 &0.0043 & 0.0049 &0.0755 & 0.0813 &0.2558 & 0.3037 &0.6680 & 0.7122  & 0.0381	&0.0656                                                \\ 
                                                                                    &                                                   & \multicolumn{1}{|l|}{RankIQA\cite{liu2017rankiqa}} &0.0400 & 0.0933 &0.0372 & 0.0523 &0.0301 & 0.0384 &0.0039 & 0.0045 &0.0846 & 0.0915 &0.2437 & 0.2969 &0.6584 & 0.7039  & 0.0385	& 0.0640                                      \\
\hline 
    \end{tabular}}
    \end{center}
    \vspace{-5mm}
    \end{table*}
    

%% file: tab/ijbc1v1.tex
\begin{table*}[ht!]
\center
\caption{Verification performance for each exit of \earlyexit (C)/(T) on the IJB-C (1:1 mixed verification) \cite{ijbc}.}
\vspace{-3mm}
\label{tab:ijbc_ver}
\resizebox{0.99\linewidth}{!}{%
\begin{tabular}{|cc|ccc|ccc|ccc|ccc|ccc|ccc|}
\hline
\multicolumn{2}{|c|}{\multirow{3}{*}{Quality Estimation}} &
\multicolumn{18}{c|}{1:1 mixed Verification: TAR (\%) at} \\ \cline{3-20}
& & \multicolumn{3}{c|}{ArcFace \cite{deng2019arcface}} & \multicolumn{3}{c|}{ElasticFace\cite{elasticface}} & \multicolumn{3}{c|}{MagFace \cite{MagFace}} & \multicolumn{3}{c|}{CurricularFace \cite{curricularFace}} & \multicolumn{3}{c|}{SwinFace \cite{Swinface}} & \multicolumn{3}{c|}{TransFace \cite{Dan_2023_TransFace}} \\
& & {FAR=1e-6} & {FAR=1e-5} & {FAR=1e-4} & {FAR=1e-6} & {FAR=1e-5} & {FAR=1e-4} & {FAR=1e-6} & {FAR=1e-5} & {FAR=1e-4} & {FAR=1e-6} & {FAR=1e-5} & {FAR=1e-4}& {FAR=1e-6} & {FAR=1e-5} & {FAR=1e-4}& {FAR=1e-6} & {FAR=1e-5} & {FAR=1e-4}\\ \hline \hline
\multicolumn{2}{|l|}{No Quality-based Aggregation} & 89.85 & 94.47  & 96.28  & 89.15 & 94.54 & 96.49 & 85.67  & 93.08 & 96.65  & 90.46 & 94.89 & 96.58 & 90.82 & 94.93 & 96.73 &86.75 & 93.87 & 96.45    \\ \hline
\multirow{14}{*}{ViT-FIQA (C)}

& \multicolumn{1}{|l|}{Exit 1} &  85.71 &  93.19 &  95.80 &  85.36 &  93.43 &  96.05 &  81.45 &  90.77 &  95.07 &  86.66 &  93.77 &  96.27 &  88.76 &  92.95 &  95.55 &  83.13 &  95.44 &  97.00\\ 
& \multicolumn{1}{|l|}{Exit 2} &  87.27 &  93.55 &  96.02 &  86.33 &  93.97 &  96.24 &  82.66 &  91.76 &  95.39 &  88.00 &  94.33 &  96.33 &  89.29 &  93.39 &  95.70 &  83.83 &  95.67 &  97.08\\ 
& \multicolumn{1}{|l|}{Exit 3} &  87.91 &  93.88 &  96.10 &  86.75 &  94.18 &  96.27 &  83.54 &  92.22 &  95.51 &  88.05 &  94.46 &  96.38 &  89.45 &  93.53 &  95.69 &  84.57 &  95.75 &  97.09\\ 
& \multicolumn{1}{|l|}{Exit 4} &  87.90 &  93.99 &  96.23 &  87.22 &  94.29 &  96.36 &  84.03 &  92.55 &  95.60 &  88.78 &  94.68 &  96.46 &  89.63 &  93.66 &  95.75 &  84.69 &  95.82 &  97.10\\
\hline

\hline
\end{tabular}
}
\end{table*}